\documentclass[times, review, 10pt]{elsarticle}

\usepackage{amssymb}
\usepackage{amsmath}

\newcommand{\ie}{\emph{i.e.}}
\newcommand{\eg}{\emph{e.g.}}
\newcommand{\cf}{\emph{cf.}}

\usepackage{graphicx}
\usepackage{booktabs}
\usepackage{amssymb}
\usepackage{amsmath}
\usepackage{fontawesome5}
\usepackage{multirow}
\usepackage{colortbl}
\usepackage{subcaption}
\usepackage{bbm}
\usepackage{adjustbox}
\usepackage{textcomp}
\usepackage[table]{xcolor}

\usepackage{xcolor}
\usepackage[hidelinks]{hyperref}

\usepackage{geometry}
\makeatletter
\newcommand{\logfontinfo}[1]{%
  \typeout{#1: font size = \f@size pt, baselineskip = \the\baselineskip}%
}
\makeatother

\begin{document}

\begin{frontmatter}



\title{Vertical Fusion: Condensing Internal Representations for Robust ViT Classification}  


\author[1]{Francesco {Di Salvo}}
\ead{francesco.di-salvo@uni-bamberg.de}

\author[1]{Shyam Nandan Rai}

\author[2]{Hamed Damirchi}

\author[2]{Ignacio Meza {De la Jara}}

\author[1]{Sebastian Doerrich}

\author[1]{Marco Lents}

\author[1]{Christian Ledig}

\affiliation[1]{
    organization={University of Bamberg},
    city={Bamberg},
    state={Bavaria},
    country={Germany}
}

\affiliation[2]{
    organization={AIML, The University of Adelaide},
    city={Adelaide},
    state={South Australia},
    country={Australia}
}


\begin{abstract}
Despite exposing rich intermediate representations, Vision Transformers (ViTs) are almost exclusively utilized as black-box feature extractors, where only the last layer is considered for downstream tasks. We challenge this convention by introducing the notion of \textit{recoverability}: the capacity of intermediate representations to correct last-layer failures. By evaluating independent classification probes at every model depth across $16$ datasets, we observe that intermediate probes correctly classify $18\%$ to $76\%$ of samples that the last-layer probe misclassifies. We show that these gains are not primarily driven by predictive diversity, but by a redundancy-correctness correspondence, where the internal hierarchy acts as a series of stable, redundant probes of a shared discriminative signal. While established \emph{horizontal} ensemble strategies (\ie, across multiple models) can improve performance, they incur high computational cost and ignore this \emph{vertical} signal within a single model. To bridge this gap, we propose \texttt{VFusion}, a principled vertical aggregation strategy employing a learnable mapping into a low-dimensional latent space that synthesizes features across the internal ViT hierarchy. \texttt{VFusion} substantially outperforms established aggregation baselines in both in-distribution and out-of-distribution settings, notably closing $45\%$ of the accuracy gap between the best individual layer and a theoretical oracle performance. Our gains consistently generalize across model sizes and pre-training regimes, confirming that \texttt{VFusion} offers a robust and efficient alternative to horizontal ensemble methods. The code is available at \href{https://github.com/francescodisalvo05/vit-vertical-fusion}{\textcolor{blue}{github.com/francescodisalvo05/vit-vertical-fusion}}.
\end{abstract}



\begin{keyword}
 Vertical Ensembling \sep 
 Internal Representations \sep 
 Recoverability \sep
 ViT
\end{keyword}

\end{frontmatter}




\section{Introduction}
\label{sec:intro}

Vision Transformers (ViTs) have achieved remarkable performance across a wide range of computer vision tasks~\citep{dosovitskiy2021an,ranftl2021vision,oquab2024dinov}, enabled by large-scale pre-training. Despite these advances, their reliability in real-world deployment remains a critical concern \citep{teterwak2025is}. In safety-critical domains such as healthcare \citep{imam2025robustness} and autonomous driving \citep{kumar2023object}, robustness to distribution shifts and inherent sample difficulty is fundamental. Current paradigms in uncertainty quantification \citep{he2025survey}, calibration \citep{minderer2021revisiting,pinto2022impartial}, and robustness evaluation \citep{gulrajani2021in, hendrycks2019robustness, pinto2022impartial} share a common implicit assumption: the last ViT layer provides the most informative representation for downstream applications. Consequently, predictive signals are almost exclusively derived from last-layer features. While effective, this reliance on the last-layer representation assumes the model has optimally retained and condensed all task-relevant information at the last layer, a design choice that remains largely undiscussed. Unlike the clear hierarchical abstraction found in convolutional neural networks, ViTs maintain a more uniform representational profile across depth \citep{raghu2021vision}. Nevertheless, downstream applications still treat them as naive feature extractors, ignoring the potential of their rich internal representations. \newline

Recent works in out-of-distribution (OOD) detection across vision~\citep{wei2025xmahalanobis}, language \citep{jelenic2024outofdistribution}, and vision-language models \citep{imezadelajara2025mysteries} suggest that this assumption may be suboptimal, as intermediate layers often encode complementary information. However, the utility of these internal representations has been primarily studied for binary classification problems (\ie, identifying OOD samples). This leaves a gap in existing work: whether \textit{corrective} or \textit{complementary} signals embedded in intermediate representations can be used to improve general multi-class classification performance. While prior work on convolutional neural networks suggests that intermediate layers can outperform the last layer under distribution shift, these findings do not consistently extend to Vision Transformers~\citep{uselis2025intermediate}. 

Due to the high representational redundancy inherent in ViTs \citep{imezadelajara2025mysteries, raghu2021vision}, extracting a relevant corrective signal across intermediate representations is non-trivial. This raises a fundamental set of questions: does an \textit{internal} corrective signal exist in intermediate ViT layers that can recover last-layer failures? If so, does this signal emerge from predictive diversity or a different representational phenomenon? \newline

Motivated by these questions, we investigate complementarity within a single frozen Vision Transformer, using lightweight layer-wise probes. We further introduce the notion of \textit{recoverability}, defined as the capacity of intermediate probes to correctly classify samples misclassified by the last-layer probe. Across $16$ diverse classification datasets covering small-scale, fine-grained, and texture-dominated settings, we find substantial internal corrective capacity: $18\%$--$76\%$ of last-layer failures are recoverable by at least one intermediate probe (\cf~Table~\ref{tab:recovery_rate}), indicating that discriminative information often persists outside the final representation. In Section~\ref{sec:layerwise_analysis}, we show that this gain is not driven by predictive diversity, but by a redundancy-correctness correspondence (\cf~Figure~\ref{fig:entropy_cosinedist}), where layers act as stable, slightly perturbed probes of a shared discriminative signal. \newline

Building upon these insights, we propose \texttt{VFusion} (Figure~\ref{fig:method_overview}), a principled aggregation strategy that employs a learnable encoder to suppress task-irrelevant information and condenses the internal hierarchy into a robust global token. Unlike traditional ensembles that are computationally expensive and rely on horizontal diversity across different models~\citep{rodriguezopazo2025synergy}, \texttt{VFusion} performs a ``vertical'' synthesis, exploiting the redundancy of ViT layers to identify and amplify stable, discriminative signals. Moreover, horizontal ensembling presupposes access to several backbones. While feasible in general vision, this assumption becomes restrictive in specialized domains such as medical image analysis, where such a diverse set of pre-trained models is often unavailable. Through an extensive evaluation across a total of $21$ classification datasets, we demonstrate that \texttt{VFusion} substantially outperforms standard aggregation methods in both in-distribution (ID, $16$ datasets) and out-of-distribution (OOD, $5$ datasets) settings. 

Our contributions can be summarized as follows: 

\begin{itemize}
	\item We introduce and quantify the concept of \textit{recoverability} in Vision Transformers, demonstrating that intermediate layers can correct between $18\%$ and $76\%$ of last-layer misclassifications. 
    \item We empirically show that this corrective signal is not driven by layer-wise diversity. Instead, robustness arises from isolating stable discriminative signals that persist across the redundant internal feature hierarchy.
	\item We propose \texttt{VFusion}, a lightweight fusion head that outperforms established aggregation baselines in both ID and OOD classification benchmarks.
    \item We provide an empirical layer-selection guideline: fusing deeper contiguous layers preserves fusion benefits, while strided subsampling is less effective.
    \item We validate \texttt{VFusion} through a broad set of ablation studies, showing that its gains persist across backbone families, model scales, and latent-dimensionality choices, while improving latent-space separability and remaining efficient.

\end{itemize}

\begin{table}[h!]
	\centering
	\scriptsize
	\setlength{\tabcolsep}{1.5pt}
    \caption{\label{tab:recovery_rate}Number of samples correctly classified by layer-wise probes \texttt{L1}-\texttt{L11} but misclassified by the last-layer probe \texttt{L12}, using a DINOv2 ViT-B backbone. We report the number of last-layer errors (\#\texttt{L12}Wrong) and samples misclassified by all layers (\#AllWrong), enabling computation of the theoretical \textit{recovery rate}, defined as the fraction of last-layer errors corrected by at least one intermediate layer. Across datasets, intermediate probes recover between $18\%$ and $76\%$ of last-layer errors. This trend is generally more pronounced for datasets with fewer classes (\textit{e.g.}, $<100$). Counts are averaged over three seeds. Note that per-layer recovery counts are not additive across layers, since multiple layers may recover the same sample.}
    \begin{tabular}{lcccccccccccccccc}
	\toprule
	Dataset & 
		\textbf{C10} & \textbf{C100} & \textbf{CUB} & \textbf{Cal} & \textbf{Cars} & \textbf{DTD} & \textbf{ESAT} & \textbf{FGVC} & \textbf{Flow} & \textbf{Food} & \textbf{GTS} & \textbf{IN1k} & \textbf{MN} & \textbf{PC} & \textbf{Pet} & \textbf{STL} \\
	\#Test & 10000 & 10000 & 5794 & 8677 & 8041 & 1880 & 27000 & 3333 & 6149 & 25250 & 12630 & 50000 & 10000 & 32768 & 3669 & 8000 \\
	\#Classes & 10 & 100 & 200 & 101 & 196 & 47 & 10 & 100 & 102 & 101 & 43 & 1000 & 10 & 2 & 37 & 10 \\
	\midrule
	\texttt{L1} & \cellcolor{green!39}17 & \cellcolor{green!10}13 & \cellcolor{green!10}4 & 0 & \cellcolor{green!10}12 & \cellcolor{green!10}9 & \cellcolor{green!50}101 & \cellcolor{green!10}6 & \cellcolor{green!10}1 & \cellcolor{green!10}21 & \cellcolor{green!10}173 & \cellcolor{green!10}7 & \cellcolor{green!14}9 & \cellcolor{green!100}1794 & \cellcolor{green!10}3 & \cellcolor{green!100}7 \\
	\texttt{L2} & \cellcolor{green!39}17 & \cellcolor{green!10}7 & \cellcolor{green!10}9 & \cellcolor{green!16}4 & \cellcolor{green!10}7 & \cellcolor{green!10}10 & \cellcolor{green!81}164 & \cellcolor{green!10}5 & 0 & \cellcolor{green!10}30 & \cellcolor{green!10}193 & \cellcolor{green!10}7 & \cellcolor{green!28}18 & \cellcolor{green!100}1671 & \cellcolor{green!10}2 & \cellcolor{green!100}8 \\
	\texttt{L3} & \cellcolor{green!37}16 & \cellcolor{green!10}8 & \cellcolor{green!10}1 & \cellcolor{green!16}4 & \cellcolor{green!10}3 & \cellcolor{green!10}9 & \cellcolor{green!76}155 & \cellcolor{green!10}4 & 0 & \cellcolor{green!10}15 & \cellcolor{green!10}196 & \cellcolor{green!10}15 & \cellcolor{green!34}22 & \cellcolor{green!100}1704 & \cellcolor{green!10}4 & \cellcolor{green!100}7 \\
	\texttt{L4} & \cellcolor{green!62}27 & \cellcolor{green!10}19 & 0 & \cellcolor{green!16}4 & \cellcolor{green!10}13 & \cellcolor{green!11}25 & \cellcolor{green!99}201 & \cellcolor{green!10}12 & 0 & \cellcolor{green!10}32 & \cellcolor{green!10}174 & \cellcolor{green!10}7 & \cellcolor{green!45}29 & \cellcolor{green!100}1777 & \cellcolor{green!10}1 & \cellcolor{green!100}8 \\
	\texttt{L5} & \cellcolor{green!69}30 & \cellcolor{green!10}7 & \cellcolor{green!10}6 & \cellcolor{green!20}5 & \cellcolor{green!10}6 & \cellcolor{green!14}31 & \cellcolor{green!100}209 & \cellcolor{green!10}8 & 0 & \cellcolor{green!10}31 & \cellcolor{green!13}264 & \cellcolor{green!10}7 & \cellcolor{green!65}42 & \cellcolor{green!100}1641 & \cellcolor{green!11}9 & \cellcolor{green!100}7 \\
	\texttt{L6} & \cellcolor{green!74}32 & \cellcolor{green!13}71 & \cellcolor{green!10}6 & \cellcolor{green!24}6 & \cellcolor{green!10}13 & \cellcolor{green!22}48 & \cellcolor{green!100}221 & \cellcolor{green!10}10 & 0 & \cellcolor{green!10}41 & \cellcolor{green!19}384 & \cellcolor{green!10}15 & \cellcolor{green!100}66 & \cellcolor{green!100}1448 & \cellcolor{green!11}9 & \cellcolor{green!100}9 \\
	\texttt{L7} & \cellcolor{green!74}32 & \cellcolor{green!22}118 & \cellcolor{green!10}19 & \cellcolor{green!64}16 & \cellcolor{green!10}10 & \cellcolor{green!27}60 & \cellcolor{green!100}212 & \cellcolor{green!10}21 & 0 & \cellcolor{green!10}68 & \cellcolor{green!21}435 & \cellcolor{green!10}62 & \cellcolor{green!100}87 & \cellcolor{green!100}1454 & \cellcolor{green!20}16 & \cellcolor{green!100}8 \\
	\texttt{L8} & \cellcolor{green!95}41 & \cellcolor{green!25}134 & \cellcolor{green!10}23 & \cellcolor{green!68}17 & \cellcolor{green!10}15 & \cellcolor{green!27}60 & \cellcolor{green!100}222 & \cellcolor{green!10}16 & 0 & \cellcolor{green!10}110 & \cellcolor{green!23}481 & \cellcolor{green!10}89 & \cellcolor{green!100}113 & \cellcolor{green!100}1427 & \cellcolor{green!21}17 & \cellcolor{green!100}9 \\
	\texttt{L9} & \cellcolor{green!95}41 & \cellcolor{green!30}159 & \cellcolor{green!13}54 & \cellcolor{green!84}21 & \cellcolor{green!10}34 & \cellcolor{green!37}81 & \cellcolor{green!100}221 & \cellcolor{green!10}41 & \cellcolor{green!10}1 & \cellcolor{green!14}211 & \cellcolor{green!27}558 & \cellcolor{green!10}246 & \cellcolor{green!100}116 & \cellcolor{green!100}1325 & \cellcolor{green!36}29 & \cellcolor{green!100}12 \\
	\texttt{L10} & \cellcolor{green!79}34 & \cellcolor{green!35}185 & \cellcolor{green!25}105 & \cellcolor{green!96}24 & \cellcolor{green!10}53 & \cellcolor{green!36}80 & \cellcolor{green!90}182 & \cellcolor{green!21}121 & \cellcolor{green!10}1 & \cellcolor{green!21}309 & \cellcolor{green!30}615 & \cellcolor{green!10}748 & \cellcolor{green!100}102 & \cellcolor{green!100}1286 & \cellcolor{green!40}32 & \cellcolor{green!100}12 \\
	\texttt{L11} & \cellcolor{green!76}33 & \cellcolor{green!38}203 & \cellcolor{green!37}156 & \cellcolor{green!56}14 & \cellcolor{green!28}271 & \cellcolor{green!33}72 & \cellcolor{green!82}167 & \cellcolor{green!45}250 & \cellcolor{green!10}1 & \cellcolor{green!31}447 & \cellcolor{green!42}851 & \cellcolor{green!15}1356 & \cellcolor{green!100}77 & \cellcolor{green!83}1070 & \cellcolor{green!39}31 & \cellcolor{green!100}9 \\
	\midrule
	\rowcolor{gray!15} \#\texttt{L12}Wrong & 132 & 948 & 659 & 59 & 1273 & 385 & 651 & 913 & 20 & 2208 & 3519 & 10971 & 268 & 4202 & 147 & 29 \\
	\rowcolor{gray!15} \#AllWrong & 43 & 528 & 415 & 25 & 945 & 218 & 202 & 555 & 16 & 1407 & 2018 & 9007 & 64 & 1286 & 79 & 7 \\
	\midrule
	\rowcolor{gray!15} \textit{Recov. rate} & 67\% & 44\% & 37\% & 58\% & 26\% & 43\% & 69\% & 39\% & 20\% & 36\% & 43\% & 18\% & 76\% & 69\% & 46\% & 76\% \\
	\bottomrule
	\end{tabular}
\end{table}


\section{Related work}

\subsection{Vision foundation models} 

Vision foundation models trained via large-scale pre-training objectives have become a dominant paradigm for general-purpose feature extraction \citep{dosovitskiy2021an,radford2021learning,oquab2024dinov}. These models are increasingly deployed as frozen backbones \citep{nakata2022revisiting,disalvo2025an,zhu2024label}, shifting the challenge of robust downstream performance from architectural design to the effective extraction and combination of their latent representations. Despite their success, model robustness depends heavily on the alignment between pre-training data and downstream benchmarks, rather than architecture alone \citep{teterwak2025is, mayilvahanan2025in}. However, these assessments rely mainly on final-layer features, overlooking the robust, task-relevant information preserved in earlier representations. While standard practice ignores the internal hierarchy, ViTs exhibit intermediate features that may encode complementary signals. We analyze and exploit this representational hierarchy for downstream classification.

\subsection{Utilizing intermediate representations}

Neural networks are traditionally viewed as learning hierarchical representations, moving from general patterns to task-specific features. Consequently, downstream applications have primarily relied on the final layer for classification, out-of-distribution detection \citep{hendrycks2017a, sun2022knnood}, and robustness analysis \citep{hendrycks2019robustness}. However, while this hierarchy is pronounced in convolutional networks, Vision Transformers exhibit a more uniform representational profile with significant redundancy across depth \citep{raghu2021vision}. This lack of sharp semantic transitions suggests that discriminative features are distributed throughout the hierarchy, inherently limiting the performance of approaches that rely on selecting a single ``optimal'' layer, regardless of how that layer is identified. \newline

Recent efforts to exploit these distributed signals have emerged primarily in out-of-distribution detection. These include strategies such as adaptive importance weighting \citep{wei2025xmahalanobis}, optimal layer selection \citep{imezadelajara2025mysteries}, learning layer-specific weights for softmax aggregation in fake image detection \citep{koutlis2024leveraging}, or characterizing the transition smoothness between intermediate representations \citep{jelenic2024outofdistribution}. While effective for binary discrimination, these methods rarely extend to the more complex requirements of multi-class classification. In this regime, Uselis et al.~\cite{uselis2025intermediate} demonstrate that intermediate probes can improve robustness, yet they report limited gains for ViTs and select layers using validation data drawn from the target distribution, a setting that is often impractical in real-world deployments. Nonetheless, we argue that these limited gains stem from the uniform nature of ViT representations, where selection-based approaches fail to capture the collective evidence available across layers. In contrast, we propose a principled aggregation strategy for combining intermediate ViT representations that exploits cross-layer information without requiring access to target distributions.

\subsection{Information fusion} 

Combining information from multiple predictors is a long-standing strategy for improving downstream performance. Classical ensemble methods \citep{dietterich2000ensemble, freund1997decision} and deep ensembles \citep{lakshminarayanan2017simple, fort2019deep} take advantage of the diversity across models to reduce variance. Related approaches aggregate model parameters rather than predictions, such as model merging \citep{wortsman2022model, ainsworth2023git} and model stitching \citep{lenc2015understanding, bansal2021revisiting}, demonstrating that meaningful information can be combined post-hoc. Beyond fixed schemes, learning-based ensembles learn to combine predictions from multiple predictors. Super Learner (SL) learns aggregation weights in logit space using only model outputs~\citep{ju2018relative}, whereas the Neural Logit Controller (NLC) performs input-conditional logit fusion by predicting aggregation weights from input features~\citep{rodriguezopazo2025synergy}. While effective, these \emph{horizontal} fusion strategies require running multiple models and benefit from the complementarity of independently trained base predictors. In contrast to ensembling across independently trained models, aggregating representations within the internal hierarchy of a single network (\ie, ``vertical fusion'') introduces distinct challenges. \newline

Intermediate layers of a Vision Transformer are highly correlated and exhibit subtler forms of complementarity, making it harder to reliably learn aggregation weights in logit space. 
Furthermore, in CLIP-style models, intermediate representations cannot be meaningfully classified in zero-shot settings because the projection into the shared vision-text space is learned only at the final layer \citep{rodriguezopazo2025synergy}. Applying the final-layer projection to earlier layers can lead to a substantial performance degradation. Motivated by these limitations, we focus on feature-level fusion within a single frozen vision backbone. Our goal is to develop an aggregation strategy that operates directly on intermediate representations and exploits their redundant yet complementary structure for downstream classification.


\section{Layer-wise analysis}
\subsection{Notation}

Let $\{(x_i, y_i)\}_{i=1}^N$ denote the training dataset of $N$ image-label pairs, with $x_i \in \mathcal{X}$ and $y_i \in \mathcal{Y}$. We consider a frozen vision backbone $\Phi$ composed of $L$ layers, used as a feature extractor. For notational simplicity, we omit the sample index $i$ when it is clear from context. Given a training image $x$, the backbone outputs a sequence of intermediate representations $\Phi(x) = \bigl(h^{(1)}, h^{(2)}, \ldots, h^{(L)}\bigr)$, where $h^{(\ell)} \in \mathbb{R}^d$ denotes the layer-$\ell$ \texttt{[CLS]} token ($d$ is the embedding dimension), used as a compact global descriptor of layer $\ell$. Collecting all \texttt{[CLS]} tokens over the training set yields a feature tensor $\mathcal{H} \in \mathbb{R}^{N \times L \times d}$, exposing the full representational hierarchy across depth. In contrast to standard practice, which typically relies only on the final layer, our goal is to exploit information \textit{across} layers. 

\subsection{Recoverability analysis}
\label{sec:layerwise_analysis}

Before introducing our fusion mechanism, we first examine whether intermediate layers contain information that can compensate for final-layer failures. Using independently trained layer-wise probes, we quantify both the amount and the structure of this corrective signal. This analysis motivates treating the ViT depth hierarchy as a source of recoverable information rather than as redundant intermediate computation.

\begin{figure}[h]
    \centering
    \includegraphics[width=1.0\linewidth]{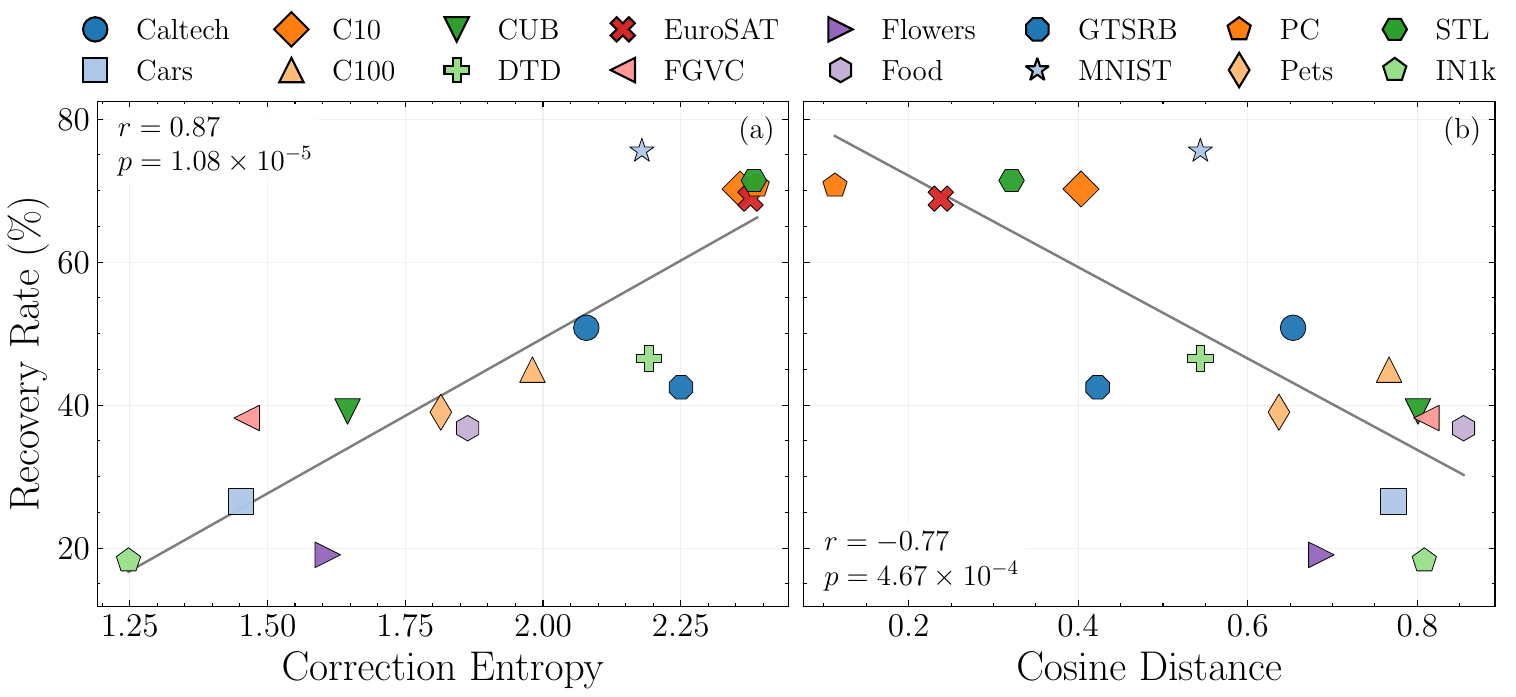}
    \caption{\label{fig:entropy_cosinedist}We measure \emph{correction entropy} $H_{\mathrm{corr}}$, defined as the entropy of the conditional correction probabilities $c_\ell = P(\ell \text{ correct} \mid L \text{ wrong})$, which quantifies how distributed correction capacity is across intermediate layers $\ell$. Higher entropy indicates that multiple layers contribute complementary boundary-level corrections. The recovery rate $(\%)$, \textit{i.e.}, fraction of last-layer $(L)$ errors recoverable by intermediate layers, correlates strongly with correction entropy ($r=0.87$), but negatively with cosine distance $D_{\mathrm{cos}}$ between logits computed on jointly misclassified samples ($r=-0.77$). These results show that recoverability can be better explained by conditional correction capacity across depth rather than layer-wise disagreement.}
\end{figure}

\subsubsection{Recoverability} 
To understand whether intermediate layers contain recoverable signal beyond the final representation, we probe each layer independently using an MLP. Let $a^{(\ell)}$ denote the accuracy at an intermediate layer $\ell$, and $a^{(L)}$ the accuracy of the final layer. As an upper bound, we define \textit{oracle accuracy} by marking a sample correct if any layer predicts the correct label:
\begin{equation}
a^{\mathrm{oracle}} = \frac{1}{N} \sum_{i=1}^{N} \mathbbm{1} \!\left(\exists \ell \in \{1,\dots,L\} : \hat{y}_i^{(\ell)} = y_i\right).
\end{equation}
Next, we define \emph{recoverability} as the fraction of final-layer errors corrected by at least one intermediate layer. Formally, we quantify it via the \emph{recovery rate}:
\begin{equation}
   \mathrm{Recovery~Rate} = \frac{a^{\mathrm{oracle}} - a^{(L)}}{100 - a^{(L)}}. 
\end{equation}

This normalization makes recovery rates comparable across datasets with different baseline accuracies. A high recovery rate indicates that information missed by the final representation remains accessible in intermediate layers.

\subsubsection{Distributed correction capacity} 
To characterize how correction capacity is distributed across depth, we consider the conditional correction probability $c_\ell$ for each intermediate layer $\ell \in \{1,\dots,L-1\}$:
\begin{equation}
c_\ell = P\big(\hat{y}^{(\ell)} = y \,\big|\, \hat{y}^{(L)} \neq y\big),
\end{equation}
which measures how often layer $\ell$ is correct on examples misclassified by the final layer. Normalizing by the sum of these probabilities over layers yields a distribution over correction mass, whose dispersion we quantify via the \emph{correction entropy}: 
\begin{equation}
\label{eq:correction_entropy}
H_{\mathrm{corr}} = - \sum_{\ell} \tilde{c}_\ell \log \tilde{c}_\ell,
\qquad
\tilde{c}_\ell = \frac{c_\ell}{\sum_k c_k}.
\end{equation}

Low correction entropy indicates that corrections are concentrated in a few layers, whereas high entropy indicates broadly distributed correction capacity across depth. 

\subsubsection{Predictive diversity}
While correction entropy $H_{\mathrm{corr}}$ captures how correction capacity is distributed across layers, it does not determine whether this capacity reflects ensemble-like diversity (high disagreement) or a shared decision boundary (low disagreement).

To disambiguate these mechanisms, we define the \emph{geometric disagreement} $D_{\mathrm{cos}}(p,q)$ as the expected cosine distance between logit vectors $u_p$ and $u_q$, restricted to samples that both layers misclassify:
\begin{equation}
D_{\mathrm{cos}}(p,q) 
= 1 - \mathbb{E}\!\left[
\frac{ u_p^\top u_q }{ \|u_p\| \, \|u_q\| }
\;\middle|\;
\hat{y}^{(p)} \neq y \wedge \hat{y}^{(q)} \neq y
\right].
\end{equation}
For each dataset, we aggregate this quantity by averaging $D_{\mathrm{cos}}(p,q)$ over all layer pairs. A low value indicates that even when two layers fail, on average, they do so in a consistent, non-stochastic manner. We further evaluate disagreement using Jensen-Shannon divergence in Appendix A.1.

\subsubsection{Key insights}
Across $16$ classification datasets, we quantify \emph{recoverability} (\cf~Table~\ref{tab:recovery_rate}), and analyze its relationship with \emph{correction entropy} and \emph{geometric disagreement} (\cf~Figure~\ref{fig:entropy_cosinedist}). We observe three trends: (i) intermediate probes recover $18\%$--$76\%$ of samples misclassified by the final layer, revealing substantial corrective capacity outside the final layer; (ii) \emph{recoverability} varies substantially across datasets, indicating that the usefulness of intermediate layers depends on dataset characteristics; (iii) \emph{recoverability} correlates strongly with correction entropy but negatively with cosine distance, suggesting that correction arises from distributed capacity across depth rather than from logit disagreement. Together, these findings point to a redundancy-correctness correspondence, where layers behave as correlated yet non-identical probes (\cf~Appendix A.2), motivating \texttt{VFusion}, our vertical ensembling strategy.

\section{\texttt{VFusion}}

\begin{figure}
\centering
\begin{subfigure}{.65\textwidth}
  \centering
  \includegraphics[width=1\linewidth]{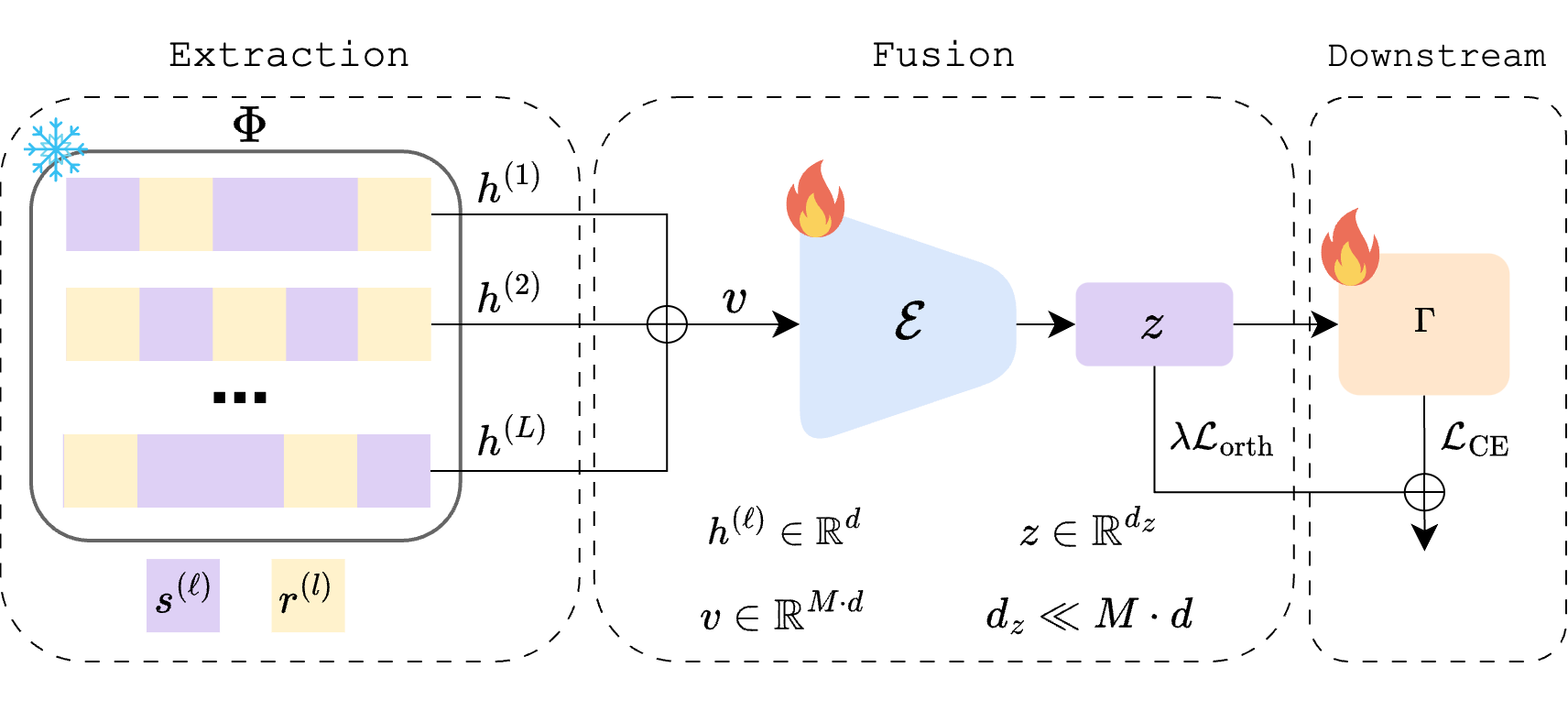}
\end{subfigure}%
\begin{subfigure}{.40\textwidth}
  \centering
  \includegraphics[width=0.8\linewidth]{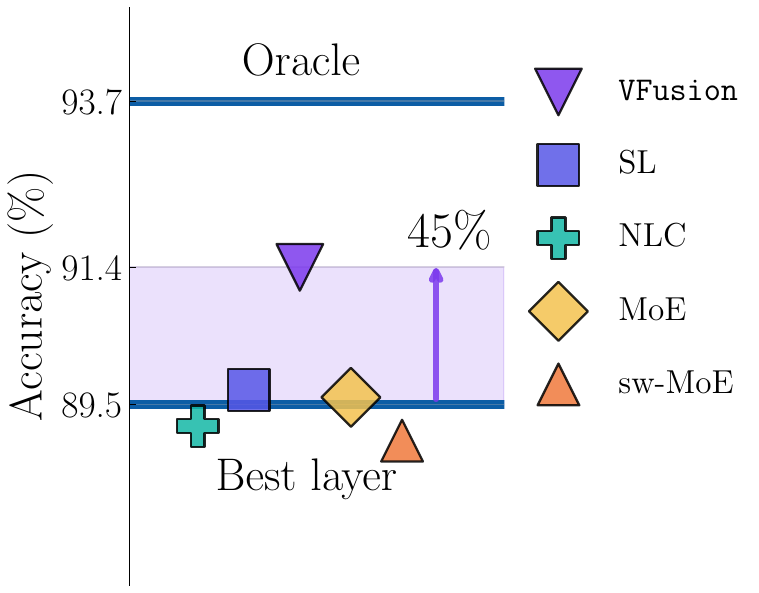}
  \vspace{5mm}
\end{subfigure}
\caption{\label{fig:method_overview}
\textbf{Left:} Illustration of \texttt{VFusion}. A frozen ViT backbone $\Phi$ extracts layer representations $\{h^{(\ell)}\}_{\ell=1}^L$, with $h^{(\ell)}=s^{(\ell)}+r^{(\ell)}$ (task-relevant + task-irrelevant). A subset of $M$ layers is selected and concatenated into $v=[h^{(\ell)}]_{\ell\in\mathcal{S}}\in\mathbb{R}^{M \cdot d}$, which is compressed by a learnable fusion encoder $\mathcal{E}$ into a low-dimensional latent $z\in\mathbb{R}^{d_z}$ ($d_z \ll M \cdot d$). A lightweight head $\Gamma$ predicts class logits from $z$ using cross-entropy loss $\mathcal{L}_{\mathrm{CE}}$, while a decorrelation regularizer $\mathcal{L}_{\mathrm{orth}}$ with weight $\lambda$ encourages decorrelated latent dimensions.
\textbf{Right:} Average accuracy across $16$ datasets. The headroom is the gap between the best-layer probe (\ie,~``Best layer'') and the theoretical oracle that marks a sample correct if \emph{any} layer predicts correctly.}
\end{figure}

The recoverability analysis shows that intermediate ViT layers contain corrective signals distributed across depth, but not in the form of independent ensemble diversity. This motivates \texttt{VFusion}: a lightweight mechanism that treats the layer hierarchy as a redundant yet informative representation trace and compresses selected frozen features into a compact task-specific latent token. Figure~\ref{fig:method_overview} illustrates the proposed framework, which consists of three components. \textit{Layer selection} chooses a subset of layers from the frozen backbone. \textit{Layer fusion} concatenates their representations and compresses them into a single global token via a learnable fusion encoder. \textit{Supervised training} jointly learns the fusion encoder and a lightweight classifier head while keeping the backbone parameters fixed. We describe each component below.

\subsection{Layer selection}

The vertical scope of the fusion manifold is defined by selecting a subset of layers $\mathcal{S} \subseteq \{1, \dots, L\}$ of size $M$. We consider three selection strategies to balance computational cost and corrective signal diversity: (i) \emph{holistic}, utilizing the full depth ($M=L$) to preserve the complete information trace; (ii) \emph{last-half} ($\mathcal{S} = \{L/2, \dots, L\}$) to focus on the deepest part of the network, and (iii) \emph{strided} ($\mathcal{S} = \{2, 4, 6, \dots\}$) (up to $L$) to reduce dimensionality while maintaining coverage across depth.

\subsection{Layer fusion}
\label{sec:layer_fusion}
Given a selected subset of layers $\mathcal{S}$, our objective is to aggregate their representations into a single \emph{global token} that consolidates complementary information distributed across depth. Our layer-wise analysis (\cf~Table~\ref{tab:recovery_rate}, Figure~\ref{fig:entropy_cosinedist}) shows that intermediate layers frequently correct mistakes made by the final layer, with recoverability strongly correlated with correction entropy.  
Importantly, this recoverability is not explained by geometric disagreement between logits, suggesting that complementarity arises from distributed conditional correction capacity rather than from purely divergent predictions. These findings motivate aggregation mechanisms that consolidate cross-layer corrective signals instead of selecting a single ``best'' layer.

\subsubsection{Cross-layer shared structure}

We interpret each layer representation $h^{(\ell)}$ as containing a task-relevant component $s^{(\ell)} \in \mathbb{R}^d$ and a task-irrelevant component $r^{(\ell)} \in \mathbb{R}^d$: $h^{(\ell)} = s^{(\ell)} + r^{(\ell)}$, 
where $\ell \in \mathcal{S}$ represents the respective layer and $d$ the embedding dimension. Under this model, multiple layers provide distinct but potentially overlapping task-relevant information. \newline

Concatenating the selected features yields a single vector
\begin{equation}
v = \big[\, h^{(\ell)} \,\big]_{\ell \in \mathcal{S}} \in \mathbb{R}^{M \cdot d},
\end{equation}
which enables a fusion model to explicitly access all relevant information, even if a specific piece of it was available in intermediate layers before being transformed into more complex but task-irrelevant information.

\subsubsection{Fusion encoder}
We introduce a learnable fusion encoder
\begin{equation}
\mathcal{E} : \mathbb{R}^{M\cdot d} \rightarrow \mathbb{R}^{d_z}, 
\quad 
z = \mathcal{E}(v), 
\quad 
z \in \mathbb{R}^{d_z},
\end{equation}
which maps the concatenated multi-layer feature vector $v$ into a low-dimensional latent $z$ with $d_z \ll M \cdot d$, and applies \textit{LayerNorm} to produce an approximately standard-normalized latent. By imposing this low-dimensional representation, the encoder acts as a noise-suppression filter that discards layer-specific residuals $r^{(\ell)}$ to isolate the task-relevant signal $s$ shared across the representational hierarchy. In practice, $\mathcal{E}$ is implemented as a multi-layer perceptron with non-linear activations, enabling flexible interactions across layers and allowing the fusion encoder to learn non-trivial cross-depth dependencies. Under this formulation, layer fusion becomes a supervised cross-layer dimensionality reduction problem: the encoder learns how to align and compress complementary evidence distributed across intermediate representations into a single global token suitable for downstream prediction. This avoids committing to any single layer and instead leverages the distributed correction capacity observed empirically in Table~\ref{tab:recovery_rate}.

\subsection{Supervised training}
\label{sec:Supervised_training}

Given the fused latent representation $z = \mathcal{E}(v) \in \mathbb{R}^{d_z}$, we attach a lightweight classifier $\Gamma : \mathbb{R}^{d_z} \rightarrow \mathbb{R}^{|\mathcal{Y}|}$ that maps the compact global token to class logits $\hat{\mathbf{y}} = \Gamma(z)$. The fusion encoder $\mathcal{E}$ and classifier $\Gamma$ are trained jointly under supervised classification loss, while the backbone $\Phi$ remains frozen throughout.

\subsubsection{Latent orthogonalization} 
Inspired by redundancy-reduction objectives \cite{zbontar2021barlow,bardes2022vicreg}, we mitigate redundancy in the fused latent representation by introducing a decorrelation regularizer that penalizes linear dependencies across dimensions. While the low-dimensional projection imposes a strong inductive bias for information compression, it does not explicitly prevent different latent dimensions from encoding redundant features. Given a mini-batch (size $B$) of encoded latent representations $\mathbf{Z} \in \mathbb{R}^{B \times d_z}$, we compute the empirical covariance matrix as:
\begin{equation}
\mathbf{C} = \frac{1}{B-1} (\mathbf{Z} - \bar{\mathbf{Z}})^\top (\mathbf{Z} - \bar{\mathbf{Z}}),
\end{equation}
where $\bar{\mathbf{Z}}$ denotes the batch mean. To ensure scale invariance across latent dimensions, we normalize $\mathbf{C}$ into the correlation matrix $\mathbf{R} = \mathbf{D}^{-1/2} \mathbf{C} \mathbf{D}^{-1/2}$, 
where $\mathbf{D} = \mathrm{diag}(\mathbf{C})$. The orthogonalization loss penalizes off-diagonal correlations as:
\begin{equation}
\mathcal{L}_{\mathrm{orth}}
=
\frac{1}{d_z(d_z - 1)}
\sum_{i \ne j} \mathbf{R}_{ij}^2.
\end{equation}

Minimizing $\mathcal{L}_{\mathrm{orth}}$ encourages different latent dimensions to encode complementary, decorrelated components of the fused representation, reducing unnecessary duplication within the global token. 

\subsubsection{Training objective} 
We train the fusion encoder and classifier end-to-end on a combination of the standard classification cross-entropy loss $\mathcal{L}_{\mathrm{CE}}$ and orthogonalization loss $\mathcal{L}_{\mathrm{orth}}$: 
\begin{equation}
\mathcal{L}_{\mathrm{total}}=\mathcal{L}_{\mathrm{CE}}+\lambda \mathcal{L}_{\mathrm{orth}},
\end{equation}
where $\lambda$ controls the strength of the decorrelation constraint.

\section{Experiments}

We evaluate \texttt{VFusion} along two complementary axes. We first assess vertical fusion for standard classification across $16$ diverse benchmarks (Section~\ref{sec:vertical_fusion_classification}), and subsequently evaluate its robustness under covariate and subpopulation shifts (Section~\ref{sec:exp_robustness}).

\subsection{Experimental details}

\subsubsection{Datasets}

We evaluate our method and established baselines on a diverse suite of widely adopted computer vision benchmarks spanning multiple domains and levels of granularity. Specifically, we consider CIFAR-10 and CIFAR-100 \citep{krizhevsky2009learning} (C10, C100), Caltech-101 \citep{fei2007learning} (Cal), Stanford Cars \citep{krause20133d} (Cars), CUB \citep{wah2011caltech}, DTD \citep{cimpoi2014describing}, EuroSAT \citep{helber2019eurosat} (ESAT), FGVC-Aircraft \citep{maji13fine-grained} (FGVC), Oxford Flowers \citep{Nilsback08} (Flow), Food-101 \citep{bossard14} (Food), GTSRB \citep{Houben-IJCNN-2013} (GTS), MNIST \citep{lecun2010mnist} (MN), PCAM \citep{pcam} (PC), Oxford-IIIT Pets \citep{parkhi2012cats} (Pets), STL-10 \citep{coates2011analysis} (STL), and ImageNet-1k \citep{imagenet} (IN1k). Furthermore, to evaluate the robustness performance of all methods, we consider datasets covering two types of distribution shifts: covariate shifts, evaluated using CIFAR-10C, CIFAR-100C, and ImageNet200-C \citep{hendrycks2019robustness} (C-10C, C-100C, IN-200C), and subpopulation shifts, assessed by CelebA \citep{liu2015deep} (celebrity face images with facial attribute annotations) and Waterbirds \citep{Sagawa2020Distributionally} (bird images with spurious background correlations).

\subsubsection{Baselines}
We compare \texttt{VFusion} against several layer-aggregation baselines, treating each intermediate layer representation as an independent expert. \textit{Non-parametric} methods combine layer-wise predictions without learning additional parameters: probability averaging (Average), majority voting (Majority), and selecting the best-performing individual layer (Best layer). \textit{Parametric} methods learn combinations of layer-wise outputs. Super Learner (SL)~\citep{ju2018relative} learns a fixed linear combination of layer predictors, whereas the Neural Logit Controller (NLC)~\citep{rodriguezopazo2025synergy} performs input-dependent calibration by predicting temperature-scaling parameters for the layer experts. \textit{Feature-based} routing methods operate on the concatenated feature representation, including sparse Mixture of Experts (MoE)~\citep{shazeer2017outrageously} and Switch-MoE (sw-MoE)~\citep{fedus2022switch}. For a fair comparison, MoE and sw-MoE receive the same concatenated features as \texttt{VFusion}. MoE uses top-$k$ routing ($k=4$), while sw-MoE is the hard-routing case ($k=1$).

\subsubsection{Implementation details}
\label{sec:implementation_details} We define all classification heads $\Gamma$ as MLPs with hidden dimension $128$, optimized with Adam under cross-entropy loss. We train with batch size $512$ for up to $20{,}000$ iterations, and apply early stopping based on validation loss evaluated every $500$ iterations: training stops if the loss does not decrease by at least $10^{-4}$ for $5$ consecutive evaluations, and we retain the checkpoint with the best validation loss. The backbone remains frozen throughout. For layer-wise probes, we select hyperparameters on a held-out validation set via a per-layer grid search over learning rate $\{10^{-4}, 10^{-3}, 10^{-2}, 10^{-1}\}$ and weight decay $\{0, 10^{-3}, 10^{-2}\}$. For MoE, we set the number of experts to $M$ and a learning rate of $2 \times 10^{-4}$ \citep{rodriguezopazo2025synergy}. For \texttt{VFusion}, we set the latent dimension to $d_z=256$ and the orthogonalization weight to $\lambda=2.0$. The encoder $\mathcal{E}$ is a $3$-layer MLP with widths $[2048,1024,512]$ and output dimension $256$. We train $(\mathcal{E},\Gamma)$ with learning rate $10^{-4}$ and weight decay $10^{-3}$. Unless stated otherwise, we adopt the \textit{holistic} layer-selection strategy. All results are averaged over three seed runs.

\subsection{Vertical fusion for classification}
\label{sec:vertical_fusion_classification}

As demonstrated in Table~\ref{tab:main_experiment}, \texttt{VFusion} achieves an average accuracy of $91.4\%$, representing a $1.9\%$ improvement over the best individual layer-wise probe and a $1.7\%$ margin over the strongest baseline, Super Learner (SL). Most baselines fail to substantially outperform the best layer on average. One exception is the DTD dataset, where SL outperforms \texttt{VFusion}. MoE and sw-MoE show a remarkably different behavior, with more pronounced gaps in fine-grained datasets. In contrast, \texttt{VFusion} exhibits substantial gains on challenging datasets such as GTS ($+7.2\%$), where samples suffer from varying resolutions and luminance. Similar improvements are observed in high-granularity tasks like Cars ($+4.4\%$) and FGVC ($+7.0\%$). \texttt{VFusion} also approaches or surpasses the theoretical oracle performance on Cars, Flow, IN1k, and ESAT, suggesting the method does not merely select the correct layer, but condenses complementary information across the network to form a~\textit{global token} that may exceed the predictive capacity of any single layer in isolation. \texttt{VFusion} closes $45\%$ of the headroom between the best layer and the theoretical oracle, capturing a substantial fraction of the unused corrective signal.

\begin{table}[h!]
	\centering
	\scriptsize
	\setlength{\tabcolsep}{2pt}
    \caption{\label{tab:main_experiment}Average accuracy (\%) obtained over $16$ datasets and averaged over three seed runs. \textit{Oracle} denotes the theoretical accuracy if any layer-wise probe classifies a sample correctly. \texttt{VFusion} outperforms the best individual layer by $1.9\%$ on average, yields gains of up to $7.2\%$, and approaches or surpasses the oracle on four datasets. We highlight in \textbf{bold} the best two methods for each dataset.}
	\begin{tabular}{lccccccccccccccccc}
	\toprule
 	& \multicolumn{5}{c}{\textbf{Small-scale}} & \multicolumn{8}{c}{\textbf{Fine-grained}} & \multicolumn{3}{c}{\textbf{Texture}} & \multicolumn{1}{c}{} \\
	\cmidrule(lr){2-6} \cmidrule(lr){7-14} \cmidrule(lr){15-17} \cmidrule(lr){18-18}
	 & \textbf{C10} & \textbf{C100} & \textbf{GTS} & \textbf{MN} & \textbf{STL} & \textbf{CUB} & \textbf{Cal} & \textbf{Cars} & \textbf{FGVC} & \textbf{Flow} & \textbf{Food} & \textbf{IN1k} & \textbf{Pet} & \textbf{DTD} & \textbf{ESAT} & \textbf{PC} & \textbf{Avg} \\
	\#Train & $10^{4}$ & $10^{4}$ & $10^{4}$ & $10^{4}$ & $10^{3}$ & $10^{3}$ & $10^{3}$ & $10^{3}$ & $10^{3}$ & $10^{3}$ & $10^{4}$ & $10^{6}$ & $10^{3}$ & $10^{3}$ & $10^{4}$ & $10^{5}$ & -- \\
	\#Classes & 10 & 100 & 43 & 10 & 10 & 200 & 101 & 196 & 100 & 102 & 101 & 1000 & 37 & 47 & 10 & 2 & -- \\
	\midrule
	\rowcolor{gray!15} Best layer & \textbf{98.7} & 90.5 & 72.1 & 97.3 & \textbf{99.6} & 88.6 & 99.3 & 84.2 & 72.6 & \textbf{99.7} & 91.3 & 78.1 & \textbf{96.0} & \textbf{79.5} & 97.6 & 87.2 & 89.5 \\ \midrule
	Average & 96.7 & 88.7 & 67.4 & 97.2 & 98.5 & \textbf{89.1} & 98.1 & 83.9 & 73.6 & 99.5 & 90.1 & 76.9 & 94.8 & 77.2 & 96.0 & 85.9 & 88.4 \\
	Majority & 87.4 & 69.8 & 38.5 & 88.0 & 88.0 & 55.0 & 60.6 & 12.4 & 34.4 & 84.2 & 69.0 & 51.6 & 64.6 & 63.8 & 93.1 & 84.7 & 65.3 \\
	SL & \textbf{98.7} & 90.6 & 73.1 & 97.5 & \textbf{99.6} & \textbf{89.1} & 99.3 & \textbf{84.4} & \textbf{74.1} & \textbf{99.7} & 91.2 & 77.9 & 95.8 & \textbf{80.0} & 97.4 & 87.3 & \textbf{89.7} \\
	NLC & 98.6 & 88.9 & 72.7 & 97.4 & \textbf{99.6} & 87.5 & 99.3 & 82.9 & 73.3 & 99.4 & 91.0 & 77.8 & 94.3 & 78.9 & 97.6 & 87.3 & 89.2 \\
	MoE & 98.6 & \textbf{90.7} & 74.5 & 98.5 & \textbf{99.6} & 87.6 & \textbf{99.5} & 80.1 & 71.2 & 99.5 & \textbf{92.0} & \textbf{81.6} & 95.5 & 78.3 & 99.1 & 87.1 & 89.6 \\
	sw-MoE & 98.6 & 90.3 & \textbf{76.7} & \textbf{98.8} & 99.5 & 86.1 & \textbf{99.5} & 78.2 & 70.9 & 97.0 & 91.9 & 78.3 & 94.4 & 75.6 & \textbf{99.4} & \textbf{88.1} & 89.0 \\
	\texttt{VFusion} & \textbf{98.7} & \textbf{91.6} & \textbf{79.3} & \textbf{98.9} & \textbf{99.7} & \textbf{89.9} & \textbf{99.6} & \textbf{88.6} & \textbf{79.6} & \textbf{99.7} & \textbf{92.8} & \textbf{82.0} & \textbf{96.0} & 79.3 & \textbf{99.6} & \textbf{87.5} & \textbf{91.4} \\ \midrule
	\rowcolor{gray!15} Oracle & 99.6 & 94.7 & 84.0 & 99.4 & 99.9 & 92.8 & 99.7 & 88.2 & 83.3 & 99.7 & 94.4 & 82.0 & 97.9 & 88.4 & 99.3 & 96.1 & 93.7 \\
	\bottomrule
	\end{tabular}
\end{table}

\subsection{Vertical fusion under distribution shift}
\label{sec:exp_robustness}

After establishing the effectiveness of \texttt{VFusion} for standard classification, we evaluate its robustness under distribution shift. As shown in Table~\ref{tab:robustness}, \texttt{VFusion} achieves the highest average accuracy ($84.6\%$) across all five benchmarks. On CelebA, Waterbirds, and C-10C, our method performs on par with the strongest baselines, yielding marginal but consistent gains. We attribute this to the lower label granularity of these tasks, where simple aggregation provides similar results. In contrast, on fine-grained settings such as C-100C and IN-200C, \texttt{VFusion} achieves larger improvements, surpassing the second-best method (best layer) by $1.3\%$ and $0.8\%$, respectively. Overall, these results suggest that while simpler strategies may suffice for low-complexity shifts, \textit{vertical fusion} becomes increasingly beneficial as task complexity and class granularity increase.

\begin{table}[h!]
	\centering
    \caption{\label{tab:robustness}Accuracy (\%) under distribution shift, including subpopulation shifts (CelebA, Waterbirds) and covariate shifts (C-10C, C-100C, IN-200C), averaged over three seed runs. \texttt{VFusion} exhibits the highest average performance, surpassing SL by $0.7\%$ on average, and outperforming the best layer by up to $1.3\%$ (Waterbirds, C-100C). We highlight in \textbf{bold} the best two methods for each dataset.}
    \small
	\begin{tabular}{lcccccc}
	\toprule
 	& \textbf{CelebA} & \textbf{Waterbirds} & \textbf{C-10C} & \textbf{C-100C} & \textbf{IN-200C} & \textbf{Avg} \\
	\#Train & $10^{5}$ & $10^{3}$ & $10^{4}$ & $10^{4}$ & $10^{5}$ & -- \\
	\#Classes & 2 & 2 & 10 & 100 & 200 & -- \\
	\midrule
	\rowcolor{gray!15} Best layer & 94.5 & 94.7 & \textbf{89.9} & \textbf{73.1} & \textbf{66.8} & 83.8 \\
	\midrule
	Average & 90.7 & 84.5 & 83.3 & 70.3 & 63.2 & 78.4 \\
	Majority & 89.7 & 82.5 & 68.2 & 52.7 & 36.8 & 66.0 \\
	SL & 94.6 & 95.2 & 89.8 & 73.0 & 66.7 & \textbf{83.9} \\
	NLC & 94.3 & \textbf{95.8} & 89.4 & 70.7 & 66.6 & 83.4 \\
	sw-MoE & \textbf{95.2} & 94.2 & 89.2 & 72.1 & 65.3 & 83.2 \\
	MoE & \textbf{95.0} & 94.9 & 89.2 & 71.9 & 65.4 & 83.3 \\
	\texttt{VFusion} & 94.9 & \textbf{96.0} & \textbf{89.9} & \textbf{74.4} & \textbf{67.6} & \textbf{84.6} \\
	\midrule
	\rowcolor{gray!15} Oracle & 98.0 & 98.6 & 95.6 & 81.0 & 72.8 & 89.2 \\
	\bottomrule
	\end{tabular}
\end{table}


\section{Ablation studies}

We conduct targeted ablation studies to analyze the main design choices of \texttt{VFusion}. In this section, we study layer selection strategies (Section~\ref{sec:layer_selection}), generalization across backbone families, model scales, and pre-training paradigms (Section~\ref{sec:backbones}), and horizontal aggregation across backbones (Section~\ref{sec:horizontal_fusion}). Additional analyses of representation geometry, orthogonalization, computational efficiency, model capacity, latent-dimension sensitivity, and extended layer-selection settings are reported in the Appendix. Together, these results clarify the benefits and robustness of \texttt{VFusion} across architectural and experimental settings.

\subsection{Layer selection strategies}
\label{sec:layer_selection}
We ablate the effect of layer selection on downstream classification performance, focusing on the two strongest feature-level fusion methods, \texttt{VFusion} and MoE. 

To ensure that differences between strategies remain non-trivial, we restrict the evaluation to $9$ datasets where performance is not saturated, \ie, with accuracy $<95\%$. All experiments use DINOv2 ViT-B as the backbone. \newline 

Table~\ref{tab:layer_selection} shows the ablation results. Although MoE achieves substantially lower overall accuracy than \texttt{VFusion}, it varies little across selection strategies. This weak sensitivity is consistent with routing that concentrates on a small subset of experts (\ie, a form of expert collapse), making MoE less sensitive to the layer pool. Likewise, \texttt{VFusion} is largely insensitive to the chosen subset: using all layers achieves the highest average accuracy, while restricting fusion to the last half incurs a minor $0.3\%$ drop, indicating that strong performance does not require carefully tuned layer selection. In contrast, the strided strategy yields a more noticeable decrease for \texttt{VFusion} (an average drop of $0.9\%$ relative to holistic), with the largest degradations on challenging benchmarks such as FGVC ($79.6\% \rightarrow 76.6\%$) and GTS ($79.3\% \rightarrow 76.8\%$). Overall, while much of the corrective signal resides in the deepest layers, earlier layers can still contribute useful evidence, and holistic selection achieves the best average performance. Therefore, we recommend using the holistic (contiguous) strategy.

\begin{table}[h]
	\centering
    \caption{\label{tab:layer_selection} Accuracy (\%) of MoE and \texttt{VFusion} under different layer-selection regimes on $9$ datasets. We report results using the DINOv2 ViT-B backbone, averaging across three seed runs. Performance is generally stable across strategies, while the holistic strategy yields the best average performance on challenging datasets such as GTS and FGVC.}
    \small
	\setlength{\tabcolsep}{5pt}
	\begin{tabular}{lcccccccccc}
	\toprule
	Selection & \textbf{C100} & \textbf{CUB} & \textbf{Cars} & \textbf{DTD} & \textbf{FGVC} & \textbf{Food} & \textbf{GTS} & \textbf{IN1k} & \textbf{PC} & \textbf{Avg} \\
	\midrule
	\rowcolor{gray!20}\multicolumn{11}{l}{MoE} \\
	\textit{Holistic} & 90.7 & 87.6 & 80.1 & 78.3 & 71.2 & 92.0 & 74.5 & 81.6 & 87.1 & 82.6 \\
	\textit{Strided} & 90.7 & 87.6 & 80.0 & 78.3 & 71.0 & 92.0 & 74.4 & 81.6 & 87.2 & 82.5 \\
	\textit{Last half} & 90.7 & 87.6 & 80.0 & 78.3 & 71.0 & 92.0 & 74.4 & 81.6 & 87.1 & 82.5 \\
	\midrule
	\rowcolor{gray!20} \multicolumn{11}{l}{\texttt{VFusion}} \\
	\textit{Holistic} & 91.6 & 89.9 & 88.6 & 79.3 & 79.6 & 92.8 & 79.3 & 82.0 & 87.5 & 85.6 \\
	\textit{Strided} & 91.4 & 89.7 & 87.6 & 79.2 & 76.6 & 92.6 & 76.8 & 82.0 & 86.6 & 84.7 \\
	\textit{Last half} & 91.5 & 89.7 & 88.4 & 79.2 & 78.4 & 93.0 & 78.5 & 82.1 & 87.4 & 85.3 \\
	\bottomrule
	\end{tabular}
\end{table}

\subsection{Generalization across backbone sizes and families}
\label{sec:backbones}

We evaluate whether the benefits of \texttt{VFusion} generalize across backbone families, model scales, and pre-training paradigms. We consider supervised IN21k ViTs \citep{dosovitskiy2021an}, self-supervised DINOv2 \citep{oquab2024dinov}, and multimodal CLIP \citep{radford2021learning} using only the vision encoder. For each family, we evaluate the available ViT scales: ViT-S, ViT-B, and ViT-L, with no official CLIP ViT-S model available. Since absolute accuracy varies across backbones, we report the average accuracy improvement ($\Delta$) relative to the best single-layer probe, isolating the benefit of vertical fusion from the base quality of the representation. \newline

\begin{figure}[h]
	\centering
    \includegraphics[width=1.0\linewidth]{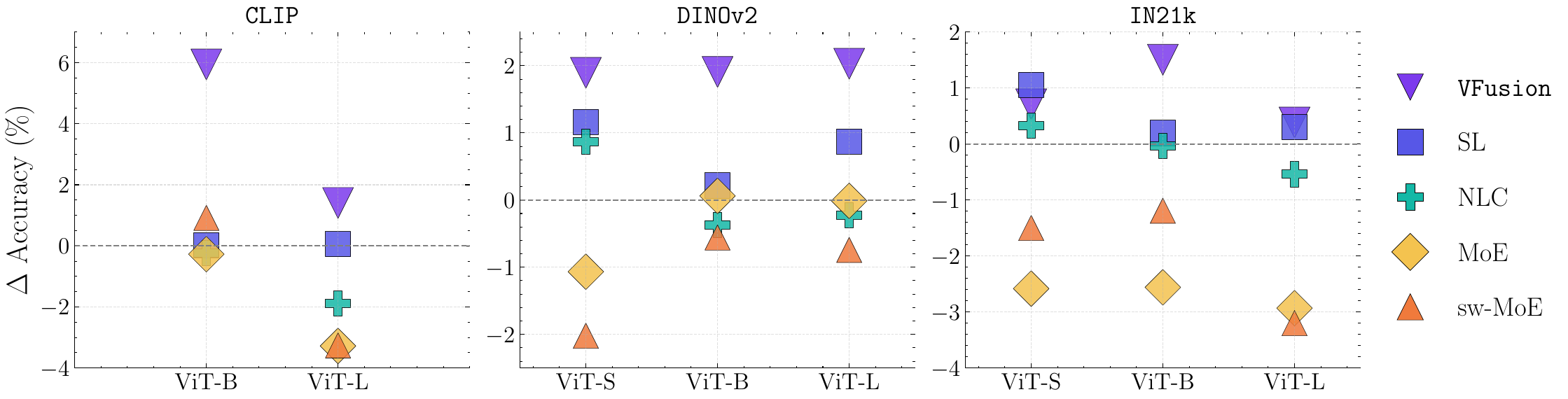}
	\caption{\label{fig:ablation_family_size} Average improvement over best-layer performance ($\Delta$ Accuracy, \%) across three ViT model scales (Small, Base, Large) and three pre-training paradigms. We evaluate supervised IN21k, self-supervised DINOv2, and CLIP (vision only) backbones over $16$ classification datasets and three seed runs. No official CLIP ViT-S model is available. \texttt{VFusion} consistently improves over the best single-layer probe, with the largest gains observed for DINOv2 and CLIP, exceeding $6\%$ on CLIP ViT-B. Conversely, reference methods exhibit generally limited improvements, mostly comparable to best-layer performance.}
\end{figure}

As shown in Figure~\ref{fig:ablation_family_size}, \texttt{VFusion} consistently improves over the best single-layer probe across backbone families and model scales. Gains are moderate for supervised IN21k backbones, reaching up to approximately $1.7\%$, and larger for self-supervised and multimodal pre-training, exceeding $6\%$ on CLIP ViT-B. In contrast, logit-level aggregation methods such as SL and NLC provide limited gains, often remaining close to best-layer performance, suggesting that the advantage of \texttt{VFusion} comes from feature-level compression of the internal hierarchy rather than prediction averaging.

\subsection{Horizontal fusion across backbones}
\label{sec:horizontal_fusion}

\subsubsection{Baselines}

We next test whether the same feature-level fusion principle extends from vertical aggregation within a model to horizontal aggregation across models. Using the three ViT-B backbones introduced above, IN21k, DINOv2, and CLIP, we treat each backbone as an expert. For logit-level baselines (Average, Majority, SL, NLC), we train an MLP probe on each backbone's last-layer features and aggregate the resulting predictions. Feature-level methods, namely sw-MoE and \texttt{HFusion}, instead operate on the concatenated last-layer features from the three backbones. Since the ensemble contains only three experts, we omit sparse MoE and report sw-MoE with $k{=}1$. Finally, \texttt{VFusion} denotes vertical fusion within the strongest single backbone, DINOv2, using holistic layer selection, whereas \texttt{HFusion} denotes horizontal fusion across the three backbones. \newline

Table~\ref{tab:horizontal} reports the horizontal-aggregation results. Compared with the vertical setting (\cf~Table~\ref{tab:main_experiment}), post-hoc aggregation benefits from cross-backbone diversity: SL and NLC improve over the best individual probe by $+1.7\%$ and $+1.5\%$, respectively. Despite this favorable setting, \texttt{VFusion} on a single DINOv2 backbone remains competitive and surpasses SL by $+0.5\%$ on average. Extending the same feature-fusion mechanism across backbones yields the best overall performance, with \texttt{HFusion} improving by a further $+0.5\%$ over \texttt{VFusion}. These results show that our fusion mechanism generalizes beyond cross-layer aggregation to heterogeneous architectures.

\begin{table}[h!]
	\centering
    \caption{\label{tab:horizontal} Accuracy (\%) on horizontal ensemble settings using three pre-trained ViT-B backbones (DINOv2, CLIP, and IN21k) across $9$ datasets. For each backbone, we train an MLP probe on the last-layer features. Logit-level baselines (Average, Majority, SL, NLC) then aggregate the resulting predictions. Feature-level methods instead operate on the concatenated last-layer representations. \texttt{VFusion} reports vertical fusion performance, \ie, within a single backbone ($\dagger$; DINOv2), while \texttt{HFusion} fuses last-layer features across backbones to learn a shared representation. Despite the increased diversity benefiting post-hoc correction methods, \texttt{VFusion} and \texttt{HFusion} achieve the best overall performance.}
    \small
	\setlength{\tabcolsep}{5pt} 
    \begin{tabular}{lcccccccccc}
	\toprule
	 & \textbf{C100} & \textbf{CUB} & \textbf{Cars} & \textbf{DTD} & \textbf{FGVC} & \textbf{Food} & \textbf{GTS} & \textbf{IN1k} & \textbf{PC} & \textbf{Avg} \\
	\#Train & $10^{4}$ & $10^{3}$ & $10^{3}$ & $10^{3}$ & $10^{3}$ & $10^{4}$ & $10^{4}$ & $10^{6}$ & $10^{5}$ & -- \\
	\#Classes & 100 & 200 & 196 & 47 & 100 & 101 & 43 & 1000 & 2 & -- \\
	\midrule
	\rowcolor{gray!15} Best layer & 90.5 & 88.6 & 84.2 & 79.5 & 72.6 & 91.3 & 78.5 & 78.1 & 87.2 & 83.4 \\ \midrule
	Average & 90.9 & 89.4 & 87.8 & \textbf{81.6} & 74.4 & 92.0 & 81.6 & 79.5 & 87.1 & 84.9 \\
	Majority & 86.6 & 87.1 & 83.0 & 78.7 & 61.0 & 90.6 & 76.9 & 73.8 & 86.6 & 80.5 \\
	SL & 91.1 & 89.4 & \textbf{89.1} & \textbf{81.5} & 75.5 & 92.5 & 80.0 & 79.4 & \textbf{87.3} & 85.1 \\
	NLC & 90.6 & 88.7 & 88.4 & 81.0 & 75.3 & 92.3 & 81.0 & 79.8 & 87.1 & 84.9 \\
	sw-MoE & 91.3 & 87.5 & 84.4 & 78.9 & 68.0 & \textbf{93.1} & \textbf{83.8} & 80.8 & 87.1 & 83.9 \\ \midrule
	\texttt{VFusion}$^{\dagger}$ & \textbf{91.6} & \textbf{89.9} & 88.6 & 79.3 & \textbf{79.6} & 92.8 & 79.3 & \textbf{82.0} & \textbf{87.5} & \textbf{85.6} \\
	\texttt{HFusion} & \textbf{91.8} & \textbf{90.1} & \textbf{89.5} & 79.6 & \textbf{76.6} & \textbf{93.4} & \textbf{84.9} & \textbf{82.9} & 86.1 & \textbf{86.1} \\ \midrule
	\rowcolor{gray!15} Oracle & 95.0 & 93.3 & 93.4 & 88.2 & 83.2 & 95.4 & 89.3 & 85.1 & 93.9 & 90.7 \\
	\bottomrule
    \end{tabular}
\end{table}

\section{Discussion}

\subsection{Limitations} 

While \texttt{VFusion} demonstrates substantial performance gains, the fusion encoder currently requires supervision to learn a task-specific aggregation of the feature representations. Future work could explore how to learn such a latent space in unsupervised or self-supervised regimes, potentially enabling large-scale fusion pre-training without the need for labeled data. 

Furthermore, layer selection is currently based on simple, predefined strategies rather than being learned jointly with the fusion encoder. We also evaluate \texttt{VFusion} primarily in the frozen-backbone setting. Studying how vertical fusion interacts with partial or full fine-tuning remains an important direction. \newline

Lastly, we evaluate \texttt{VFusion} only for supervised classification. Future work can adapt vertical fusion to other computer vision tasks, including dense prediction (\eg, segmentation).

\subsection{Conclusions} 

In this work, we provide quantitative evidence of the substantial predictive value residing within the hidden representations of Vision Transformers. We propose \texttt{VFusion}, a vertical aggregation strategy that outperforms established baselines by effectively condensing the high redundancy of intermediate layers. Our experiments demonstrate that latent redundancy is a beneficial asset rather than a limitation, as \texttt{VFusion} learns a compact global token that generally outperforms any single layer and established aggregation strategies, even surpassing the theoretical layer-wise oracle in several benchmarks. These gains are particularly pronounced in fine-grained and degraded image tasks, where the fusion module acts as a robust filter for subtle semantic signals. Furthermore, we demonstrate that our method is stable across various latent sizes and regularization parameters while generalizing across different backbone scales and pre-training regimes. \newline 

Beyond vertical aggregation, our horizontal fusion results show that the same feature-level compression principle also extends to heterogeneous backbones, yielding additional gains when complementary representations from different pre-trained models are available. We also find that performance generally benefits from exploiting information throughout the entire hierarchy, rather than from a selected subset. Crucially, these improvements are achieved with minimal computational overhead, providing a lightweight and efficient mechanism for maximizing the utility of pre-trained, off-the-shelf models.



\section*{CRediT authorship contribution statement}

\textbf{Francesco Di Salvo:} Conceptualization, methodology, software, investigation, validation, writing -- original draft.
\textbf{Shyam Nandan Rai:} Conceptualization, methodology, investigation, writing -- review and editing.
\textbf{Hamed Damirchi:} Conceptualization.
\textbf{Ignacio Meza De la Jara:} Conceptualization.
\textbf{Sebastian Doerrich:} Writing -- review and editing.
\textbf{Marco Lents:} Writing -- review and editing.
\textbf{Christian Ledig:} Supervision, conceptualization, writing -- review and editing, funding acquisition.

\section*{Declaration of competing interest}

The authors declare that they have no known competing financial interests or personal relationships that could have influenced the work reported in this paper.

\section*{Acknowledgments}

This study was funded through the Hightech Agenda Bayern (HTA) of the Free State of Bavaria, Germany.

\section*{Data availability}

The datasets used in this study are publicly available benchmark datasets. All data sources are referenced and described in the manuscript. 

\section*{Declaration of generative AI and AI-assisted technologies in the manuscript preparation process}

The core ideas, research design, and initial drafting of this manuscript were conducted entirely by the human authors. During the refinement phase, the authors utilized GPT models from OpenAI strictly to assist with language polishing, proofreading, and improving overall text readability. The final manuscript was fully reviewed, verified, and approved by the authors, who take full responsibility for the content of the published article.

\bibliographystyle{elsarticle-num} 
\bibliography{bibliography}





\clearpage

\setcounter{page}{1}

\appendix

\begin{center}
{\LARGE Vertical Fusion: Condensing Internal Representations for Robust ViT Classification} \newline
\end{center}

\section*{Appendix}

This document reports additional analyses complementing the main manuscript. \ref{sec:ablation_layerwise} extends the layer-wise analysis by examining predictive diversity and localized recovery. \ref{sec:ablation_geometry} studies the geometry of the fused latent representation and the effect of orthogonalization. \ref{sec:efficiency_capacity_sensitivity} evaluates inference efficiency, capacity control, and sensitivity to the latent dimension. \ref{sec:appendix_layer_selection} further analyzes layer-selection strategies on larger backbones and for Super Learner.

\section{Layer-wise analysis}
\label{sec:ablation_layerwise}

\setcounter{table}{0}
\setcounter{figure}{0}
\renewcommand{\thetable}{A\arabic{table}}
\renewcommand{\thefigure}{A\arabic{figure}}

\subsection{Predictive diversity via Jensen-Shannon (JS) divergence}
\label{sec:appendix_jensen}

\begin{figure}
    \centering
    \includegraphics[width=1.0\linewidth]{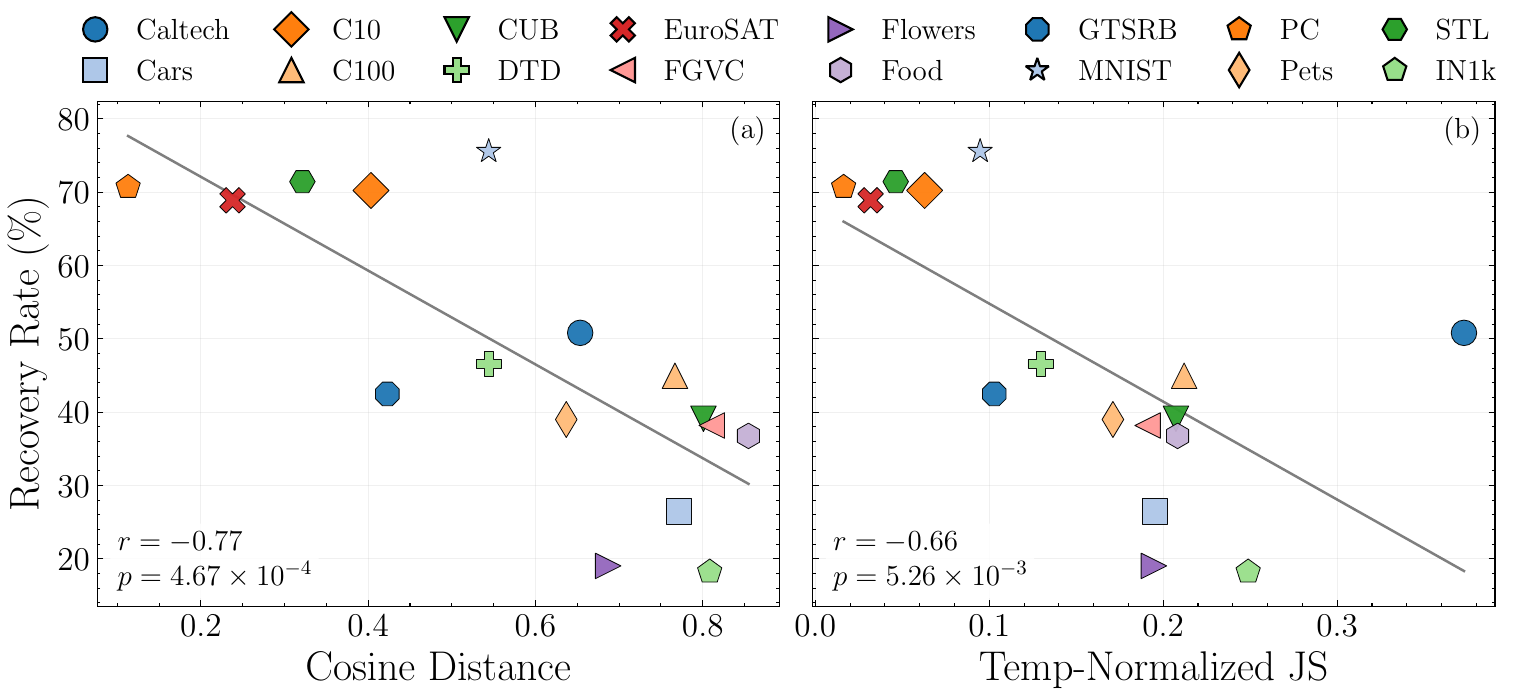}
    \caption{\label{fig:diversity_js}Recovery rate (\%) versus average pairwise layer disagreement computed on jointly misclassified samples. Disagreement is measured by (a) cosine distance between layer logits and (b) temperature-normalized Jensen-Shannon (JS) divergence between the corresponding softmax distributions. In both cases, higher recoverability is associated with lower disagreement, suggesting that oracle gains are not driven by ensemble-like diversity.} 
\end{figure}

This ablation complements the notion of predictive diversity in Section~3.2 of the main manuscript. Besides cosine distance between logit vectors, we report the temperature-normalized Jensen-Shannon (JS) divergence between the corresponding softmax distributions. Unlike cosine distance, which can remain small when two layers produce similarly oriented logits but differ in calibration, JS captures differences in the full predictive distribution. To account for logit-scale differences across layers, we apply a layer-wise temperature normalization prior to the softmax. Disagreement is computed on samples that are jointly misclassified by the two layers and then averaged over layer pairs. Overall, Figure~\ref{fig:diversity_js} confirms the same trend under both metrics.

\subsection{Localized recovery}
\label{sec:appendix_localized_recovery}

A natural question regarding layer aggregation in ViTs is whether the high representational redundancy renders additional layers superfluous. However, we argue that this redundancy is a key strength rather than a limitation. If layers act as correlated yet non-identical probes, they might correct mistakes made by their neighbors. \newline

Table~\ref{tab:local_recovery} quantifies this effect by reporting the number of samples uniquely recovered by each layer relative to its neighbors. As observed, while layers are functionally similar, they exhibit significant local complementarity. We quantify this by counting samples that are correctly classified by a specific layer $\ell$ while being misclassified by its immediate neighbors ($\neg c_{\ell-1} \land c_\ell \land \neg c_{\ell+1}$). Our findings reveal that (i) every layer across the hierarchy, from the early blocks to the final layer, uniquely recovers a subset of samples that its neighbors fail to identify. For instance, in the EuroSAT (ESAT) dataset, \texttt{L1} uniquely corrects $1,154$ samples that \texttt{L2} misses. (ii) The distribution of these local recoveries varies by domain and label granularity. In fine-grained tasks like FGVC-Aircraft (FGVC) and Stanford Cars (Cars), unique successes are concentrated in the deeper layers ($\texttt{L10}-\texttt{L12}$), whereas in texture-based or simpler tasks like EuroSAT (ESAT), PCAM (PC), or MNIST (MN), the early layers contribute more. This local complementarity demonstrates that while the layers are redundant in aggregate, they are not functionally identical.

\begin{table}[h!]
	\centering
	\scriptsize
    \caption{\label{tab:local_recovery}Each entry reports the number of test samples that are correctly classified at layer $\ell$ (\ie, \texttt{L1}-\texttt{L12}) while being misclassified by its neighboring layers. For intermediate layers, we count samples satisfying $\neg b_{\ell-1} \land b_\ell \land \neg b_{\ell+1}$. For the boundary layers, we use one-sided variants: \texttt{L1} counts $b_1 \land \neg b_2$, and layer~$L$ (\texttt{L12}) counts $\neg b_{L-1} \land b_L$. Cells are shaded with a green gradient proportional to the respective count, normalized by the number of samples correctly classified at the final layer ($\#$\texttt{L12}-C).}
	\setlength{\tabcolsep}{1.7pt}
    \begin{tabular}{lcccccccccccccccc}
	\toprule
	Dataset & \textbf{C10} & \textbf{C100} & \textbf{CUB} & \textbf{Cal} & \textbf{Cars} & \textbf{DTD} & \textbf{ESAT} & \textbf{FGVC} & \textbf{Flow} & \textbf{Food} & \textbf{GTS} & \textbf{IN1k} & \textbf{MN} & \textbf{PC} & \textbf{Pet} & \textbf{STL} \\
	\#Test & 10000 & 10000 & 5794 & 8677 & 8041 & 1880 & 27000 & 3333 & 6149 & 25250 & 12630 & 50000 & 10000 & 32768 & 3669 & 8000 \\
	\#Classes & 10 & 100 & 200 & 101 & 196 & 47 & 10 & 100 & 102 & 101 & 43 & 1000 & 10 & 2 & 37 & 10 \\
	\midrule
	$\texttt{L1}$ & \cellcolor{green!5}514 & \cellcolor{green!1}100 & \cellcolor{green!0}26 & \cellcolor{green!0}24 & \cellcolor{green!0}53 & \cellcolor{green!1}18 & \cellcolor{green!4}1154 & \cellcolor{green!0}22 & \cellcolor{green!1}62 & \cellcolor{green!0}167 & \cellcolor{green!1}121 & \cellcolor{green!0}50 & \cellcolor{green!0}16 & \cellcolor{green!7}2104 & \cellcolor{green!1}68 & \cellcolor{green!4}355 \\
	$\texttt{L2}$ & \cellcolor{green!4}479 & \cellcolor{green!1}100 & \cellcolor{green!0}30 & \cellcolor{green!0}18 & \cellcolor{green!0}45 & \cellcolor{green!0}10 & \cellcolor{green!3}811 & \cellcolor{green!0}22 & \cellcolor{green!0}29 & \cellcolor{green!0}167 & \cellcolor{green!0}61 & \cellcolor{green!0}50 & \cellcolor{green!2}286 & \cellcolor{green!1}366 & \cellcolor{green!0}31 & \cellcolor{green!3}268 \\
	$\texttt{L3}$ & \cellcolor{green!2}283 & \cellcolor{green!1}100 & \cellcolor{green!0}25 & \cellcolor{green!0}3 & \cellcolor{green!0}44 & \cellcolor{green!0}9 & \cellcolor{green!0}71 & \cellcolor{green!1}33 & \cellcolor{green!1}74 & \cellcolor{green!1}250 & \cellcolor{green!0}24 & \cellcolor{green!0}50 & \cellcolor{green!0}45 & \cellcolor{green!0}157 & \cellcolor{green!1}38 & \cellcolor{green!1}134 \\
	$\texttt{L4}$ & \cellcolor{green!3}319 & \cellcolor{green!1}127 & \cellcolor{green!0}28 & \cellcolor{green!0}1 & \cellcolor{green!0}37 & \cellcolor{green!2}32 & \cellcolor{green!1}329 & \cellcolor{green!1}33 & \cellcolor{green!0}7 & \cellcolor{green!1}250 & \cellcolor{green!0}38 & \cellcolor{green!0}50 & \cellcolor{green!0}33 & \cellcolor{green!1}368 & \cellcolor{green!0}32 & \cellcolor{green!4}367 \\
	$\texttt{L5}$ & \cellcolor{green!2}281 & \cellcolor{green!1}127 & \cellcolor{green!0}23 & \cellcolor{green!0}3 & \cellcolor{green!0}29 & \cellcolor{green!1}29 & \cellcolor{green!0}123 & \cellcolor{green!1}36 & \cellcolor{green!0}20 & \cellcolor{green!1}250 & \cellcolor{green!0}75 & \cellcolor{green!0}51 & \cellcolor{green!1}123 & \cellcolor{green!0}181 & \cellcolor{green!1}65 & \cellcolor{green!2}174 \\
	$\texttt{L6}$ & \cellcolor{green!1}177 & \cellcolor{green!3}319 & \cellcolor{green!0}29 & \cellcolor{green!0}10 & \cellcolor{green!0}44 & \cellcolor{green!3}57 & \cellcolor{green!0}138 & \cellcolor{green!0}19 & \cellcolor{green!1}119 & \cellcolor{green!1}317 & \cellcolor{green!2}242 & \cellcolor{green!0}60 & \cellcolor{green!1}155 & \cellcolor{green!0}239 & \cellcolor{green!1}63 & \cellcolor{green!1}98 \\
	$\texttt{L7}$ & \cellcolor{green!1}105 & \cellcolor{green!2}270 & \cellcolor{green!0}41 & \cellcolor{green!0}46 & \cellcolor{green!0}19 & \cellcolor{green!2}31 & \cellcolor{green!0}137 & \cellcolor{green!1}39 & \cellcolor{green!1}112 & \cellcolor{green!1}434 & \cellcolor{green!1}133 & \cellcolor{green!0}282 & \cellcolor{green!0}78 & \cellcolor{green!0}216 & \cellcolor{green!1}57 & \cellcolor{green!0}66 \\
	$\texttt{L8}$ & \cellcolor{green!0}48 & \cellcolor{green!1}154 & \cellcolor{green!0}22 & \cellcolor{green!0}22 & \cellcolor{green!0}18 & \cellcolor{green!1}28 & \cellcolor{green!0}56 & \cellcolor{green!0}17 & \cellcolor{green!0}54 & \cellcolor{green!1}382 & \cellcolor{green!1}168 & \cellcolor{green!0}310 & \cellcolor{green!0}43 & \cellcolor{green!0}203 & \cellcolor{green!1}44 & \cellcolor{green!0}23 \\
	$\texttt{L9}$ & \cellcolor{green!0}41 & \cellcolor{green!1}176 & \cellcolor{green!1}84 & \cellcolor{green!0}30 & \cellcolor{green!1}71 & \cellcolor{green!2}31 & \cellcolor{green!0}109 & \cellcolor{green!1}42 & \cellcolor{green!1}61 & \cellcolor{green!1}458 & \cellcolor{green!3}280 & \cellcolor{green!1}577 & \cellcolor{green!0}35 & \cellcolor{green!0}209 & \cellcolor{green!0}32 & \cellcolor{green!0}13 \\
	$\texttt{L10}$ & \cellcolor{green!0}18 & \cellcolor{green!1}112 & \cellcolor{green!1}88 & \cellcolor{green!0}22 & \cellcolor{green!0}59 & \cellcolor{green!1}25 & \cellcolor{green!0}48 & \cellcolor{green!4}105 & \cellcolor{green!0}6 & \cellcolor{green!1}349 & \cellcolor{green!2}207 & \cellcolor{green!2}967 & \cellcolor{green!0}30 & \cellcolor{green!1}314 & \cellcolor{green!0}22 & \cellcolor{green!0}5 \\
	$\texttt{L11}$ & \cellcolor{green!0}16 & \cellcolor{green!1}93 & \cellcolor{green!2}111 & \cellcolor{green!0}3 & \cellcolor{green!3}245 & \cellcolor{green!1}23 & \cellcolor{green!0}46 & \cellcolor{green!8}196 & \cellcolor{green!0}1 & \cellcolor{green!1}255 & \cellcolor{green!4}441 & \cellcolor{green!2}989 & \cellcolor{green!0}17 & \cellcolor{green!0}247 & \cellcolor{green!0}14 & \cellcolor{green!0}3 \\
	$\texttt{L12}$ & \cellcolor{green!1}128 & \cellcolor{green!8}807 & \cellcolor{green!5}271 & \cellcolor{green!1}143 & \cellcolor{green!28}1898 & \cellcolor{green!10}156 & \cellcolor{green!2}597 & \cellcolor{green!15}381 & \cellcolor{green!0}40 & \cellcolor{green!13}3114 & \cellcolor{green!14}1332 & \cellcolor{green!27}10667 & \cellcolor{green!1}146 & \cellcolor{green!4}1191 & \cellcolor{green!2}91 & \cellcolor{green!0}46 \\
	\midrule
	\rowcolor{gray!15} \#$\texttt{L12}$-C & 9867 & 9052 & 5135 & 8618 & 6767 & 1494 & 26348 & 2420 & 6129 & 23042 & 9111 & 39028 & 9732 & 28565 & 3521 & 7971 \\
	\bottomrule
	\end{tabular}
\end{table}


\section{Analysis of the fused manifold}
\label{sec:ablation_geometry}

\setcounter{table}{0}
\setcounter{figure}{0}
\renewcommand{\thetable}{B\arabic{table}}
\renewcommand{\thefigure}{B\arabic{figure}}

This section analyzes the representation learned by \texttt{VFusion}. We evaluate class separability relative to the frozen last-layer representation and assess the effect of orthogonalization on accuracy and latent redundancy.


\subsection{Class separability of the fused latent space}

To study how \texttt{VFusion} reshapes representation geometry, we compare the frozen last-layer \texttt{[CLS]} token $h^{(L)}$ with the fused latent $z=\mathcal{E}(v)$, using DINOv2 ViT-B. Although the backbone is fixed, the supervised fusion module can reorganize redundant cross-layer information into a more separable representation. In particular, we evaluate class structure on normalized test feature embeddings using two complementary metrics. Let $f_i$ denote the normalized feature embedding of sample $i$ in the feature space being evaluated: $f_i=h_i^{(L)}$ for the last-layer representation and $f_i=z_i$ for \texttt{VFusion}'s latent. First, we compute a centroid-based \emph{separation ratio},
\begin{equation}
\mathrm{Sep}
=
\frac{
\frac{2}{C(C-1)}
\sum_{1 \leq c < c' \leq C}
\|\mu_c-\mu_{c'}\|_2
}{
\frac{1}{N}
\sum_{i=1}^{N}
\|f_i-\mu_{y_i}\|_2
},
\end{equation}

where $\mu_c$ denotes the centroid of class $c$ and $C$ is the number of classes. This metric increases when class centroids move farther apart relative to the within-class spread, providing an intuitive measure of centroid-level separation. 

Second, we compute the \emph{Fisher trace ratio},
\begin{equation}
\mathrm{Fisher}=\frac{\mathrm{tr}(S_B)}{\mathrm{tr}(S_W)},
\end{equation}
where $S_B$ and $S_W$ denote the between-class and within-class scatter matrices, respectively. While the separation ratio captures centroid-level geometry, the Fisher ratio additionally summarizes the global balance between inter-class and intra-class scatter. \newline

Table~\ref{tab:separation} shows consistent improvements across all datasets and for both separability metrics. The separation ratio increases by $1.8\times$--$17.9\times$, while the Fisher ratio increases by $2.8\times$--$191.5\times$ relative to the last-layer embedding. Since these metrics capture complementary aspects of class geometry, their agreement indicates that \texttt{VFusion} both increases inter-class separation and reduces relative within-class scatter. These improvements also extend to fine-grained datasets such as Cars, FGVC, Food, and Pets, suggesting that the fused latent space preserves discriminative structure useful for subtle class distinctions. Overall, \texttt{VFusion} not only improves predictive accuracy, but reorganizes the redundant cross-layer hierarchy into a substantially more class-structured latent manifold.

\subsection{Effect of orthogonalization}
\label{sec:ablation_orthogonalization}
To reduce redundancy in the fused manifold, we add an orthogonalization penalty to the training objective, weighted by $\lambda$. This term encourages the encoder to extract decorrelated and complementary factors from the concatenated hierarchy. We evaluate its effect by sweeping $\lambda \in \{0.2, 0.5, 0.7, 1.0, 2.0, 5.0, 10.0\}$ and compare against the unregularized baseline ($\lambda=0$). \newline

As summarized in Table~\ref{tab:orthogonalization}, the inclusion of the orthogonalization loss yields an overall improvement. 
The average gain is partially diluted by datasets whose performance is already near-saturated (\eg, $\geq 95\%$), where there is limited headroom for improvement. Nonetheless, the gains are most pronounced in fine-grained datasets, such as Cars (up to $+0.6\%$), FGVC (up to $+1.5\%$), and Food (up to $+0.8\%$), where resolving subtle inter-class differences benefits from a more diverse feature set. Importantly, the benefit is not tied to a single favorable value of $\lambda$: across the full $50\times$ sweep, the mean orthogonalized model improves over the unregularized baseline. We use $\lambda=2.0$ throughout our experiments unless stated otherwise. \newline

To confirm that our fused manifold indeed reduces redundancy, we measure a test-set proxy of $\mathcal{L}_{\mathrm{orth}}$, namely the mean squared off-diagonal correlation of the fused latent dimensions. Across all datasets, setting the orthogonalization-loss weight to $\lambda{=}2$ consistently reduces this quantity compared to $\lambda{=}0$, with a $0.25\times$ average ratio and a $0.01\text{--}0.75\times$ min--max range. This further confirms that the regularizer effectively decorrelates the fused representation.

\begin{table}[h!]
	\centering
    \scriptsize
    \caption{\label{tab:separation}
    We compare the frozen last-layer \texttt{[CLS]} embedding (\textit{Last}) with the \texttt{VFusion} latent on the test set. All features are normalized prior to metric computation. \emph{Separation ratio} denotes the centroid-based separation ratio, defined as the mean inter-class centroid distance divided by the mean intra-class distance to the class centroid (higher is better). \emph{Fisher} denotes the trace ratio $\mathrm{tr}(S_B)/\mathrm{tr}(S_W)$ computed from between- and within-class scatter matrices (higher is better). The $\times$ rows report the multiplicative improvement of \texttt{VFusion} over the last-layer representation. Across datasets, the fused latent improves separation by $1.8\times$--$17.9\times$ and Fisher separability by $2.8\times$--$191.5\times$ over the last-layer representation.}
	\setlength{\tabcolsep}{2pt} 
    \begin{tabular}{lcccccccccccccccc}
	\toprule
  	& \textbf{C10} & \textbf{C100} & \textbf{CUB} & \textbf{Cal} & \textbf{Cars} & \textbf{DTD} & \textbf{ESAT} & \textbf{FGVC} & \textbf{Flow} & \textbf{Food} & \textbf{GTS} & \textbf{IN1k} & \textbf{MN} & \textbf{PC} & \textbf{Pet} & \textbf{STL} \\
	\midrule
	\rowcolor{gray!20}\multicolumn{17}{l}{\textit{Separation ratio}} \\
	Last & 0.73 & 0.94 & 1.93 & 1.69 & 1.26 & 0.89 & 1.02 & 1.32 & 2.54 & 1.38 & 0.77 & 1.26 & 0.88 & 0.40 & 1.63 & 0.72 \\
	\texttt{VFusion} & 3.32 & 2.92 & 3.66 & 4.99 & 2.74 & 1.61 & 2.91 & 2.53 & 7.24 & 3.06 & 1.46 & 2.65 & 2.76 & 1.05 & 3.62 & 12.99 \\
	$\times$ & 4.5$\times$ & 3.1$\times$ & 1.9$\times$ & 3.0$\times$ & 2.2$\times$ & 1.8$\times$ & 2.8$\times$ & 1.9$\times$ & 2.8$\times$ & 2.2$\times$ & 1.9$\times$ & 2.1$\times$ & 3.1$\times$ & 2.6$\times$ & 2.2$\times$ & 17.9$\times$ \\ \midrule
	\rowcolor{gray!20}\multicolumn{17}{l}{\textit{Fisher}} \\
	Last & 0.24 & 0.43 & 1.79 & 1.24 & 0.80 & 0.39 & 0.50 & 0.92 & 3.00 & 0.92 & 0.24 & 0.77 & 0.39 & 0.03 & 1.25 & 0.24 \\
	\texttt{VFusion} & 4.49 & 3.26 & 5.03 & 6.80 & 3.21 & 1.14 & 3.50 & 2.64 & 20.27 & 3.54 & 0.93 & 2.89 & 3.10 & 0.27 & 5.22 & 46.12 \\
	$\times$ & 18.3$\times$ & 7.5$\times$ & 2.8$\times$ & 5.5$\times$ & 4.0$\times$ & 2.9$\times$ & 7.0$\times$ & 2.9$\times$ & 6.8$\times$ & 3.9$\times$ & 3.9$\times$ & 3.8$\times$ & 8.0$\times$ & 7.7$\times$ & 4.2$\times$ & 191.5$\times$ \\
	\bottomrule
	\end{tabular}
\end{table}

\begin{table}[h!]
	\centering
	\scriptsize
    \caption{\label{tab:orthogonalization} We compare the baseline ($\lambda=0$) against the aggregate statistics (min, mean, max) of the orthogonalized models with $\lambda \in \{0.2, 0.5, 0.7, 1.0, 2.0, 5.0, 10.0\}$. The results demonstrate that while the method is robust to the specific choice of $\lambda$, the addition of the orthogonalization constraint benefits especially fine-grained datasets like Cars ($+0.6\%$), FGVC ($+1.5\%$), and Food ($+0.8\%$).}
	\setlength{\tabcolsep}{2.4pt}
	\begin{tabular}{lccccccccccccccccc}
	\toprule
	Stat & \textbf{C10} & \textbf{C100} & \textbf{CUB} & \textbf{Cal} & \textbf{Cars} & \textbf{DTD} & \textbf{ESAT} & \textbf{FGVC} & \textbf{Flow} & \textbf{Food} & \textbf{GTS} & \textbf{IN1k} & \textbf{MN} & \textbf{PC} & \textbf{Pet} & \textbf{STL} & \textbf{Avg} \\
	\midrule
	$\lambda=0$ & 98.7 & 91.4 & 89.7 & 99.6 & 88.0 & 78.7 & 99.3 & 78.1 & 99.7 & 92.3 & 79.1 & 82.1 & 98.8 & 87.3 & 96.1 & 99.7 & 91.2 \\ \midrule
	Min & 98.7 & 91.4 & 89.8 & 99.6 & 88.1 & 78.8 & 99.5 & 75.9 & 99.7 & 92.3 & 78.9 & 81.9 & 98.9 & 86.7 & 95.8 & 99.7 & 91.0 \\
	Mean & 98.7 & 91.6 & 89.9 & 99.6 & 88.3 & 79.2 & 99.6 & 78.4 & 99.7 & 92.6 & 79.2 & 82.1 & 99.0 & 87.3 & 96.0 & 99.7 & 91.3 \\
	Max & 98.8 & 91.7 & 89.9 & 99.7 & 88.6 & 79.5 & 99.6 & 79.6 & 99.7 & 93.1 & 79.5 & 82.3 & 99.0 & 88.0 & 96.1 & 99.7 & 91.5 \\
	\bottomrule
	\end{tabular}
\end{table}


\section{Model efficiency and sensitivity } 
\label{sec:efficiency_capacity_sensitivity}

\setcounter{table}{0}
\setcounter{figure}{0}
\renewcommand{\thetable}{C\arabic{table}}
\renewcommand{\thefigure}{C\arabic{figure}}

We further report three additional analyses addressing practical aspects of \texttt{VFusion}. Specifically, we compare its inference cost against horizontal ensembling, test whether its gains are explained by increased model capacity, and assess sensitivity to the latent dimensionality $d_z$. These analyses complement the accuracy results by examining whether the proposed vertical fusion mechanism remains efficient, capacity-controlled, and robust to implementation choices.

\subsection{Computational efficiency} 

To quantify computational efficiency, we compare \texttt{VFusion} against horizontal Super Learner (H-SL), a strong ensemble baseline that aggregates predictions from three frozen backbones. This comparison is intentionally conservative: H-SL benefits from cross-backbone diversity, but requires evaluating multiple models at inference time. In contrast, \texttt{VFusion} operates within a single frozen backbone and adds only a lightweight fusion encoder and classifier head. We use CIFAR-100 as a representative benchmark for measuring end-to-end cost. \newline

Table~\ref{tab:efficiency} shows that \texttt{VFusion} substantially reduces inference cost compared to H-SL, requiring $60.5\%$ fewer GFLOPs and achieving a $2.32\times$ speedup in latency per sample. Importantly, this efficiency gain does not come at the cost of accuracy: across 9 datasets (\cf~Table~5 of the main manuscript), \texttt{VFusion} also outperforms H-SL by $+0.5\%$ on average. 

These results show that vertical fusion provides an efficient alternative to horizontal ensembling, preserving the benefits of representation aggregation while avoiding the overhead of evaluating multiple backbones.

\begin{table} [h]
    \centering
    \small
    \caption{\label{tab:efficiency}End-to-end inference efficiency on CIFAR-100. We compare \texttt{VFusion} with horizontal Super Learner (H-SL), which ensembles three frozen backbones. \texttt{VFusion} substantially reduces computational cost, requiring $60.5\%$ fewer GFLOPs and achieving a $2.32\times$ latency speedup per sample.}
    \setlength{\tabcolsep}{6.0pt}
    \begin{tabular}{lcc}
    \toprule
    \textbf{Method}  & \textbf{End-to-End (GFLOPs)} & \textbf{End-to-End Latency (ms/sample)} \\
    \midrule
    H-SL &  55.72 & 3.14 \\
    \texttt{VFusion} & \textbf{22.01 \textcolor{teal}{(-60.5\%)}} & \textbf{1.35 \textcolor{teal}{(2.32$\times$)}} \\
    \bottomrule
    \end{tabular}
\end{table}

\subsection{Capacity-controlled final-layer baseline} 

We next test whether \texttt{VFusion}'s gains come from intermediate-layer information rather than added module capacity. We introduce a capacity-controlled \textit{Final-only} baseline with the same \texttt{Enc+LN+Head} architecture and training protocol as \texttt{VFusion}, but using only the final-layer representation $h^{(L)}$ as input. 

As shown in Table~\ref{tab:final_only}, \texttt{VFusion} consistently outperforms this capacity-matched baseline on average ($91.4\%$ vs.\ $90.7\%$). The difference is small on saturated datasets, where both methods already approach ceiling performance, but becomes pronounced on less-saturated benchmarks. In particular, \texttt{VFusion} notably improves over \textit{Final-only} on GTS ($+4.3\%$), FGVC ($+4.2\%$), Cars ($+1.2\%$), and PC ($+0.8\%$). These results indicate that the gains observed in previous experiments are not merely due to a larger classifier head, but arise from fusing complementary information distributed across intermediate layers.

\begin{table}[h!]
     \centering
     \scriptsize
     \caption{\label{tab:final_only} Capacity-controlled comparison between \textit{Final-only} and \texttt{VFusion}. \textit{Final-only} uses the same \texttt{Enc+LN+Head} architecture as \texttt{VFusion}, but receives only the final-layer representation $h^{(L)}$ as input, thereby having a comparable number of trainable parameters. \texttt{VFusion} improves average accuracy by $+0.7\%$, with the largest gains on less-saturated datasets such as GTS, FGVC, Cars, and PC ($+0.8\%$--$+4.3\%$).}
	\setlength{\tabcolsep}{2pt}
     \begin{tabular}{lccccccccccccccccc}
     \toprule
      & \textbf{C10} & \textbf{C100} & \textbf{CUB} & \textbf{Cal} & \textbf{Cars} & \textbf{DTD} & \textbf{ESAT} & \textbf{FGVC} & \textbf{Flow} & \textbf{Food} & \textbf{GTS} & \textbf{IN1k} & \textbf{MN} & \textbf{PC} & \textbf{Pet} & \textbf{STL} & \textbf{Avg} \\
     \midrule

     \textit{Final-only} & \textbf{98.8} & 91.3 & 89.7 & \textbf{99.6} & 87.4 & 78.6 & 99.5 & 75.4 & \textbf{99.7} & 92.7 & 75.0 & 81.7 & 98.7 & 86.7 & \textbf{96.1} & 99.6 & 90.7 \\
     \texttt{VFusion} & 98.7 & \textbf{91.6} & \textbf{89.9} & \textbf{99.6} & \textbf{88.6} & \textbf{79.3} & \textbf{99.6} & \textbf{79.6} & \textbf{99.7} & \textbf{92.8} & \textbf{79.3} & \textbf{82.0} & \textbf{98.9} & \textbf{87.5} & 96.0 & \textbf{99.7} & \textbf{91.4} \\
     \bottomrule
     \end{tabular}
\end{table}

\subsection{Latent dimension sensitivity} 
\label{sec:ablation_latent_dimension}

Finally, we analyze the sensitivity of \texttt{VFusion} to the latent dimension $d_z$, which controls the capacity of the fusion encoder $\mathcal{E}$. Since $d_z$ defines the compression bottleneck between the concatenated layer features and the classifier head, it determines how aggressively the model must compress the internal hierarchy. 

We sweep $d_z \in \{64, 128, 256, 512\}$ while disabling the orthogonalization term ($\lambda=0$), thereby isolating the effect of latent dimensionality from the decorrelation regularizer. \newline

Table~\ref{tab:latent_dimensionality} shows that \texttt{VFusion} is robust to the choice of $d_z$. On simple datasets with near-saturated performance, accuracy remains essentially unchanged across latent dimensions. Even on more challenging datasets such as FGVC-Aircraft and DTD, the performance spread remains moderate. Across datasets, the mean $\Delta_{\mathrm{max}}$ is only $0.3\%$, indicating that the method is largely insensitive to this hyperparameter. This suggests that \texttt{VFusion} does not depend on a carefully tuned latent capacity. Rather, the redundant cross-layer hierarchy can be compressed over a broad range of bottleneck sizes. We therefore use $d_z=256$ throughout our experiments as a balanced choice between compression and signal preservation.

\begin{table}[h]
	\centering
	\scriptsize
    \caption{\label{tab:latent_dimensionality}Sensitivity of \texttt{VFusion} to the latent dimension $d_z$. We report the accuracy (\%) across $16$ datasets and three seeds, sweeping $d_z \in \{64, 128, 256, 512\}$. We highlight the \colorbox{red!20}{minimum} and \colorbox{green!20}{maximum} accuracy per dataset in red and green, respectively. We denote the per-dataset difference as $\Delta_{\mathrm{max}}$, where a lower value indicates low sensitivity to $d_z$. While optimal trade-offs are dataset-dependent, the average $\Delta_{\mathrm{max}} = 0.3\%$ confirms that \texttt{VFusion} is notably robust to the selection of latent dimensionality, likely due to the highly redundant nature of the internal representations it synthesizes.}
	\setlength{\tabcolsep}{1.5pt} 
	\begin{tabular}{lccccccccccccccccc}
	\toprule
	$d_z$ & \textbf{C10} & \textbf{C100} & \textbf{CUB} & \textbf{Cal} & \textbf{Cars} & \textbf{DTD} & \textbf{ESAT} & \textbf{FGVC} & \textbf{Flow} & \textbf{Food} & \textbf{GTS} & \textbf{IN1k} & \textbf{MN} & \textbf{PC} & \textbf{Pet} & \textbf{STL} & \textbf{Avg} \\
	\midrule
	$d_z=64$ & \cellcolor{green!20}98.7 & \cellcolor{red!20}91.4 & \cellcolor{red!20}89.5 & \cellcolor{green!20}99.6 & \cellcolor{red!20}87.1 & \cellcolor{red!20}78.3 & \cellcolor{green!20}99.6 & 77.2 & \cellcolor{green!20}99.7 & 92.3 & \cellcolor{red!20}79.1 & 82.1 & \cellcolor{red!20}98.7 & \cellcolor{red!20}87.0 & \cellcolor{green!20}96.1 & \cellcolor{green!20}99.7 & \cellcolor{red!20}91.0 \\
	$d_z=128$ & \cellcolor{green!20}98.7 & \cellcolor{green!20}91.5 & 89.6 & \cellcolor{green!20}99.6 & 87.5 & 78.7 & \cellcolor{red!20}99.2 & 78.0 & \cellcolor{green!20}99.7 & \cellcolor{green!20}92.4 & \cellcolor{red!20}79.1 & \cellcolor{green!20}82.2 & \cellcolor{green!20}98.9 & \cellcolor{green!20}87.5 & \cellcolor{green!20}96.1 & \cellcolor{green!20}99.7 & 91.1 \\
	$d_z=256$ & \cellcolor{green!20}98.7 & \cellcolor{red!20}91.4 & \cellcolor{green!20}89.7 & \cellcolor{green!20}99.6 & \cellcolor{green!20}88.0 & 78.7 & 99.3 & \cellcolor{green!20}78.1 & \cellcolor{green!20}99.7 & 92.3 & \cellcolor{red!20}79.1 & 82.1 & 98.8 & 87.3 & \cellcolor{green!20}96.1 & \cellcolor{green!20}99.7 & \cellcolor{green!20}91.2 \\
	$d_z=512$ & \cellcolor{green!20}98.7 & \cellcolor{red!20}91.4 & 89.6 & \cellcolor{green!20}99.6 & 87.7 & \cellcolor{green!20}79.0 & 99.5 & \cellcolor{red!20}77.0 & \cellcolor{red!20}99.6 & \cellcolor{red!20}92.0 & \cellcolor{green!20}79.3 & \cellcolor{red!20}81.9 & \cellcolor{green!20}98.9 & 87.1 & \cellcolor{red!20}96.0 & \cellcolor{green!20}99.7 & 91.1 \\
	\midrule
	$\Delta_{\max}$ & 0.0 & 0.1 & 0.2 & 0.0 & 0.9 & 0.7 & 0.4 & 1.1 & 0.1 & 0.4 & 0.2 & 0.3 & 0.2 & 0.5 & 0.1 & 0.0 & 0.3 \\
	\bottomrule
	\end{tabular}
\end{table}

\section{Layer selection}
\label{sec:appendix_layer_selection}

\setcounter{table}{0}
\setcounter{figure}{0}
\renewcommand{\thetable}{D\arabic{table}}
\renewcommand{\thefigure}{D\arabic{figure}}

\subsection{Effect on larger backbones}
\label{sec:layer_selection_large}

This ablation complements Section 6.1 of the main manuscript by evaluating layer selection strategies (holistic, strided, last-half) on \texttt{VFusion} and MoE with DINOv2 ViT-L, which doubles the hierarchical depth of the ViT-B model ($L=24$ vs. $L=12$). \newline 

As shown in Table~\ref{tab:layer_selection_large}, we observe trends consistent with the ViT-B backbone. MoE is essentially invariant to the layer selection, consistent with routing concentrating on a small subset of experts and thus limiting sensitivity to the available layers. 

For \texttt{VFusion}, holistic fusion remains a safe default and achieves the best (or tied-best) average performance. On ViT-L, restricting fusion to the last half (\ie, last $12$ layers) matches holistic on average, suggesting that much of the recoverable signal is concentrated in deeper layers for this backbone. However, holistic fusion never hurts and retains access to any complementary evidence present earlier in the hierarchy. In contrast, strided subsampling consistently underperforms (by $0.5\%$ on average), with the largest degradations on challenging datasets such as FGVC ($84.8\% \rightarrow 83.0\%$) and GTS ($81.7\% \rightarrow 80.5\%$). \newline

Overall, \texttt{VFusion} is robust to layer selection, but when subsampling is required, preserving contiguous depth (holistic or last-half) is preferable to striding.

\subsection{Effect on Super Learner}
\label{sec:ablation_sl}

Super Learner (SL) aggregates predictions by learning optimal weights over layer-wise logits. In the vertical setting, a natural concern is that incorporating the full hierarchy could introduce noise from shallower, less discriminative layers and thus degrade performance. To test this, we ablate SL under the three-layer selection regimes (holistic, strided, last-half) on two DINOv2 backbones: ViT-B and ViT-L. \newline 

Table~\ref{tab:super_learner_layers} shows that Super Learner is largely robust to the inclusion of shallow layers. On both DINOv2 ViT-B and ViT-L, holistic aggregation matches last-half results, indicating that incorporating the full hierarchy does not degrade performance. Strided selection yields only a minor drop (by $0.3\%$ on ViT-B, and by $0.2\%$ on ViT-L). Overall, SL does not appear sensitive to potential ``noise'' from earlier probes in the vertical setting. Furthermore, even in this logit-level aggregation setting, strided selection underperforms last-half, consistent with our observation that a useful corrective signal is locally distributed across adjacent layers.

\begin{table}[h!]
	\centering
	\small
    \caption{\label{tab:layer_selection_large} Accuracy (\%) of MoE and \texttt{VFusion} under different layer-selection regimes across $9$ datasets. We report results using the DINOv2 ViT-L backbone, averaging across three seed runs.}
	\setlength{\tabcolsep}{1.5pt}
    \begin{tabular}{lcccccccccc}
    \toprule
    Selection & \textbf{C100} & \textbf{CUB} & \textbf{Cars} & \textbf{DTD} & \textbf{FGVC} & \textbf{Food} & \textbf{GTS} & \textbf{IN1k} & \textbf{PC} & \textbf{Avg} \\
    \midrule
    \rowcolor{gray!20}\multicolumn{11}{l}{MoE} \\
    \textit{Holistic} & 92.9 & 89.8 & 82.7 & 80.0 & 74.2 & 93.7 & 72.8 & 84.8 & 87.5 & 84.3 \\
    \textit{Strided} & 92.9 & 89.7 & 82.8 & 80.0 & 74.6 & 93.7 & 72.5 & 84.8 & 87.3 & 84.3 \\
    \textit{Last half} & 93.0 & 89.7 & 82.8 & 80.1 & 74.4 & 93.7 & 72.8 & 84.8 & 87.5 & 84.3 \\
    \midrule
    \rowcolor{gray!20}\multicolumn{11}{l}{\texttt{VFusion}} \\
    \textit{Holistic} & 93.8 & 91.2 & 91.3 & 81.3 & 84.8 & 94.1 & 81.7 & 84.2 & 87.5 & 87.8 \\
    \textit{Strided} & 93.6 & 91.1 & 90.9 & 80.9 & 83.0 & 93.9 & 80.5 & 84.2 & 87.9 & 87.3 \\
    \textit{Last half} & 93.6 & 91.2 & 91.4 & 80.8 & 84.7 & 94.0 & 81.4 & 84.2 & 88.5 & 87.8 \\
    \bottomrule
    \end{tabular}
\end{table}

\begin{table}[h!]
	\centering
	\small
    \caption{\label{tab:super_learner_layers}Accuracy (\%) of Super Learner (SL) under three layer-selection strategies on DINOv2 ViT-B ($L=12$) and ViT-L ($L=24$), averaged over $9$ datasets and three seed runs.}
	\setlength{\tabcolsep}{1.5pt}
    \begin{tabular}{lcccccccccc}
    \toprule
    Selection & \textbf{C100} & \textbf{CUB} & \textbf{Cars} & \textbf{DTD} & \textbf{FGVC} & \textbf{Food} & \textbf{GTS} & \textbf{IN1k} & \textbf{PC} & \textbf{Avg} \\
    \midrule
    \rowcolor{gray!20}\multicolumn{11}{l}{Super Learner - DINOv2 ViT-B} \\
    \textit{Holistic} & 90.6 & 89.1 & 84.4 & 80.0 & 74.1 & 91.2 & 73.1 & 77.9 & 87.3 & 83.1 \\
    \textit{Last half} & 90.6 & 89.1 & 84.5 & 80.1 & 74.1 & 91.2 & 73.1 & 77.9 & 87.3 & 83.1 \\
    \textit{Strided} & 90.5 & 88.6 & 84.1 & 80.2 & 72.8 & 91.2 & 72.2 & 77.9 & 87.3 & 82.8 \\ \midrule
    \rowcolor{gray!20}\multicolumn{11}{l}{Super Learner - DINOv2 ViT-L} \\
    \textit{Holistic} & 93.2 & 90.8 & 88.9 & 81.8 & 82.4 & 93.6 & 75.1 & 80.7 & 88.6 & 86.1 \\
    \textit{Last half} & 93.1 & 90.7 & 88.9 & 81.6 & 82.5 & 93.6 & 75.1 & 80.8 & 88.7 & 86.1 \\
    \textit{Strided} & 93.0 & 90.8 & 88.7 & 81.8 & 81.8 & 93.6 & 74.8 & 80.7 & 88.3 & 85.9 \\
    \bottomrule
    \end{tabular}
\end{table}

\end{document}